\documentclass{article}

\usepackage[preprint]{neurips_2026}

\usepackage[utf8]{inputenc} 
\usepackage[T1]{fontenc}    
\usepackage[colorlinks,
            linkcolor=red,
            anchorcolor=blue,
            citecolor=green
            ]{hyperref}
\usepackage{url}            
\usepackage{booktabs}       
\usepackage{amsfonts}       
\usepackage{nicefrac}       
\usepackage{microtype}      
\usepackage{xcolor}         
\usepackage[dvipsnames]{xcolor}         
\usepackage[table]{xcolor}
\usepackage{graphicx}
\usepackage{xspace}
\usepackage{amsmath}
\usepackage{enumitem}
\usepackage{soul}
\usepackage{multirow}
\usepackage{makecell}
\usepackage{tabularx}
\usepackage{setspace}
\usepackage{utfsym}
\usepackage{pifont}
\usepackage{arydshln}
\usepackage{tcolorbox}
\usepackage{wrapfig}
\setcitestyle{square,comma,numbers}
\usepackage{colortbl}
\usepackage{algorithm}
\usepackage{algorithmic}
\usepackage{amssymb}
\usepackage{caption}

\newcommand{\eg}{\emph{e.g.,}\xspace}
\newcommand{\ie}{\emph{i.e.,}\xspace}

\definecolor{lightgrayv}{HTML}{EEEDF3} 
\definecolor{grayv}{HTML}{707070}
\definecolor{redv}{HTML}{C00000}
\definecolor{bluev}{HTML}{0070C0}

\newcommand{\baby}{\textsc{Motab}\xspace}

\title{Backtracking When It Strays: Mitigating Dual Exposure Biases in LLM Reasoning Distillation}

\author{%
  Bing Wang$^{1,2,3}$ \quad Shaotian Yan$^{3}$ \quad Chen Shen$^{3}$\thanks{Corresponding authors} \quad Kaiyuan Liu$^{4}$ \quad Sinan Fan$^{4}$ \\
  \textbf{Ximing Li}$^{1,2,6}$$^*$ \quad \textbf{Rui Miao}$^{3,5}$ \quad \textbf{Xiaosong Yuan}$^{1,2,3}$ \quad \textbf{Zhanming Shen}$^{3,4}$ \quad \textbf{Jieping Ye}$^{3}$ \\
  \small \textsuperscript{1} College of Computer Science and Technology, Jilin University \\
  \small \textsuperscript{2} Key Laboratory of Symbolic Computation and Knowledge Engineering, MoE, Jilin University \\
  \small \textsuperscript{3} Tongyi Lab, Alibaba Group \ 
  \textsuperscript{4} College of Computer Science and Technology, Zhejiang University \\
  \small \textsuperscript{5} School of Artificial Intelligence, Jilin University \
  \textsuperscript{6} RIKEN Center for Advanced Intelligence Project \\
  \small \texttt{\{wangbing1416,zjushenchen,liximing86\}@gmail.com}
}

\begin{document}

\maketitle

\begin{abstract}
  Large language models (LLMs) have achieved remarkable success in complex reasoning tasks via long chain-of-thought (CoT), yet their immense computational overhead hinders real-world deployment. LLM reasoning distillation addresses this by transferring reasoning capabilities from formidable teacher models to compact student models. However, existing distillation paradigms face a fundamental dilemma. Typical off-policy distillation strictly utilizes teacher-generated golden trajectories, suffering from an \textit{exposure bias} due to the mismatch between training distributions and student-generated inference contexts, which leads to error cascades in long CoT reasoning. To address this, on-policy distillation allows students to explore their own trajectories, but we demonstrate that it inherently introduces a reciprocal \textit{reversed exposure bias}: the teacher model also struggles to provide positive guidance when conditioned on student-generated sub-optimal contexts. To resolve this dual exposure bias problem, we propose \textit{Monitoring Trajectories and Backtracking when it strays} (\baby), a new LLM reasoning distillation pipeline. Specifically, \baby dynamically monitors the student's on-policy generation against an adaptive safety boundary. When the generation strays and exceeds this threshold, \baby backtracks to the last safe state and leverages teacher intervention to correct the course. This approach inherently tolerates minor student errors to mitigate exposure bias, while preventing sub-optimal contexts to circumvent reversed exposure bias. Extensive experiments on the \textit{LIMO-v2} and \textit{AceReason} datasets demonstrate that \baby effectively alleviates the dual exposure biases, yielding a roughly 3\% average performance improvement in reasoning tasks.
\end{abstract}

\section{Introduction} \label{sec:intro}

Recently, various \textit{large language models} (LLMs), \eg DeepSeek \citep{guo2025deepseek}, have demonstrated remarkable capabilities in complex reasoning tasks that require multi-step \textit{chain-of-thought} (CoT) \citep{wei2022chain,shao2024deepseekmath,yang2025qwen3,yuan2026differential}. 
Despite their efficacy, deploying these state-of-the-art models in real-world applications remains computationally intensive, motivating the adoption of knowledge distillation to transfer their reasoning capabilities into smaller, faster student models, namely \textit{LLM reasoning distillation} \citep{agarwal2024on,yang2025qwen3,yan2026distribution}.

Typically, most previous works follow an \textit{off-policy distillation} paradigm \citep{ho2023large,muennighoff2025s1}: a teacher model generates an extensive corpus of reasoning trajectories for a given set of complex problems; these trajectories are subsequently filtered using heuristic pipelines \citep{li2025naturalthoughts,guha2025openthoughts,jung2025prismatic} and utilized for supervised fine-tuning of the student model.
However, this off-policy approach empirically suffers from an \textit{exposure bias} issue \citep{agarwal2024on,xu2025speculative}. Specifically, during the training phase, the student model is exclusively optimized using the \textit{golden} trajectories provided by the teacher model. Conversely, during inference, the student model is exposed to its own \textit{sub-optimal} generated context. This mismatch inevitably leads to an error cascade, particularly within the context of long CoT reasoning trajectories \citep{yan2026distribution}.

To mitigate this issue, the community recently proposed an alternative paradigm, namely \textit{on-policy distillation} \citep{agarwal2024on,ko2024distillm}. Specifically, the student model generates its own reasoning trajectories, which are then guided through dense feedback provided by the teacher model, \eg token probabilities \citep{agarwal2024on,gu2024minillm,ko2025distillm}.
Although this method effectively eliminates exposure bias by exposing the student's own inference-time trajectories, in this work, we formally and empirically demonstrate that on-policy distillation also introduces a reciprocal \textit{reversed exposure bias} challenge. Analogously, when the student model generates sub-optimal on-policy trajectories as context, the teacher model often suppresses these tokens and struggles to provide positive supervisory signals \citep{fu2026revisiting}.
As experimentally evidenced by Figs.~\ref{logits} and \ref{confidence_difference}, a consistent discrepancy exists between the teacher's and student’s next-token probability distributions when conditioned on student-generated on-policy trajectories, and the divergence between the teacher's distributions under off-policy versus on-policy contexts consistently increases as the sequence length extends.

Therefore, to specifically address these \textit{dual exposure biases}, the ideal distillation dataset should \textit{cover} the student model's own sub-optimal on-policy data distribution to mitigate exposure bias, while the teacher model must provide supervision within a \textit{relatively accurate} student-generated context to circumvent reversed exposure bias, which creates an apparent dilemma. To achieve these goals, we propose a new LLM reasoning distillation method, namely \textit{Monitoring Trajectories and Backtracking when it strays} (\baby). 
Specifically, \baby consistently monitors the student model's on-policy trajectories against an adaptive safety boundary, \eg based on the entropy of teacher-generated tokens. Once a generation step exceeds this threshold, \baby backtracks to the last safe state through a credit assignment strategy, where the teacher model intervenes to steer the subsequent generation. 
Accordingly, this approach utilizes a safety boundary to tolerate sub-optimal on-policy trajectories, encouraging the model to expose its own errors and thereby mitigating the exposure bias. When the student model strays, \baby employs backtracking to identify a high-quality on-policy context, allowing the teacher model to provide guidance that effectively addresses the reversed exposure bias.

To evaluate the performance of \baby, we utilize Qwen3-32B \citep{yang2025qwen3} and gpt-oss-120b \citep{openai2025gpt} as teacher models to generate distillation data on the questions from \textit{LIMO-v2} \citep{ye2025limo} and \textit{AceReason} \citep{liu2025acereason} datasets using \baby, and perform full-parameter fine-tuning on three distinct student models. Experimental results demonstrate that our approach yields an average improvement of approximately 3\% in complex reasoning performance and effectively mitigates dual exposure biases.
\textit{Our source code is released in the repository} \url{https://github.com/wangbing1416/MOTAB}.

In summary, our contributions can be summarized as the follow three-folds:
\begin{itemize}
    \item We formally and experimentally summarize a fundamental dilemma inherent in existing off-policy and on-policy distillation methods, \ie the dual exposure biases problem.
    \item To address the biases, we introduce \baby, a new reasoning distillation framework that monitors student-generated on-policy trajectories within an adaptive safety boundary, and backtracks to the last safe point when the student model strays.
    \item Extensive experimental results demonstrate that our approach consistently enhances LLM reasoning performance and alleviates the dual exposure biases.
\end{itemize}

\section{Investigation on Dual Exposure Biases} \label{sec.2}

In this section, we formally define and empirically investigate the dual exposure biases in LLM reasoning distillation. The detailed theoretical analysis of these biases is provided in Appendix~\ref{app:theory_detailed}.

\subsection{Formal Definition of Dual Exposure Biases}

\noindent
\textbf{Task descriptions of LLM reasoning and its distillation.}
We formalize the LLM reasoning process as an auto-regressive sequence generation task. Given an initial question $\mathbf{q}$, a policy LLM $\boldsymbol{\pi}$ generates a reasoning trajectory consisting of a sequence of reasoning steps $\mathcal{Y} = (\mathbf{y}_1, \mathbf{y}_2, \dots, \mathbf{y}_L)$. At any step $l$, the reasoning context $\mathbf{s}_l$ is defined as the concatenation of the question and the previously generated prefix as $\mathbf{s}_l = [\mathbf{q}, \mathcal{Y}_{<l}]$. The policy $\boldsymbol{\pi}(\cdot \mid \mathbf{s}_l)$ defines the probability distribution over the vocabulary space for the next generation step. We denote $d^{\boldsymbol{\pi}}(\mathbf{s})$ as the marginal distribution of contexts induced by the policy $\boldsymbol{\pi}$ during free-form inference. The ultimate goal of reasoning distillation is to optimize a student policy $\boldsymbol{\pi}_S$ to accurately replicate the reasoning capabilities of a capable teacher policy $\boldsymbol{\pi}_T$.

\noindent
\textbf{Formulation of dual exposure biases.}
Existing reasoning distillation methods can be broadly categorized into off-policy and on-policy approaches. However, both paradigms inherently suffer from different forms of exposure biases:

\begin{itemize}
    \item \textbf{Exposure bias of off-policy distillation:} Typical off-policy methods follow the standard teacher-forcing objective, which minimizes the \textit{Kullback-Leibler} (KL) divergence over the teacher's context distribution $d^{\boldsymbol{\pi}_T}$:
    \begin{equation}
    \label{eq1}
        \mathcal{J}_\text{off}(\boldsymbol{\pi}_S) = \mathbb{E}_{\mathbf{s} \sim d^{\boldsymbol{\pi}_T}} 
        \big[ D_\text{KL}(\boldsymbol{\pi}_T(\cdot \mid \mathbf{s}) \parallel \boldsymbol{\pi}_S(\cdot \mid \mathbf{s})) \big].
    \end{equation}
    However, during test-time inference, the student auto-regressively generates trajectories from its own distribution $d^{\boldsymbol{\pi}_S}$. Because the training distribution mismatches the inference distribution ($d^{\boldsymbol{\pi}_T} \neq d^{\boldsymbol{\pi}_S}$), a distribution shift occurs. Once the student makes an initial error, it encounters a context $\mathbf{s} \notin \text{supp}(d^{\boldsymbol{\pi}_T})$, inevitably leading to cascading errors.

    \item \textbf{Reversed exposure bias of on-policy distillation:} To mitigate this distribution shift, on-policy methods directly sample contexts from the student's own distribution $d^{\boldsymbol{\pi}_S}$:
    \begin{equation}
    \label{eq2}
        \mathcal{J}_\text{on}(\boldsymbol{\pi}_S) = \mathbb{E}_{\mathbf{s} \sim d^{\boldsymbol{\pi}_S}} 
        \big[ D_\text{KL}(\boldsymbol{\pi}_S(\cdot \mid \mathbf{s}) \parallel \boldsymbol{\pi}_T(\cdot \mid \mathbf{s})) \big].
    \end{equation}
    However, when $\mathbf{s} \sim d^{\boldsymbol{\pi}_S}$ contains severe logical flaws due to the student's sub-optimal capabilities, this trajectory acts as an \textit{out-of-distribution} input for the teacher. Consequently, the supervision signal $\boldsymbol{\pi}_T(\cdot \mid \mathbf{s})$ always suppresses student tokens and fails to provide the positive feedback, ultimately degrading into uninformative noise.
\end{itemize}
Accordingly, existing methods struggle to simultaneously align the training with the inference-time distribution while maintaining valid, high-quality supervision, which forms a fundamental \textit{dilemma}, resulting in the \textbf{dual exposure biases}. More formulations can be seen in Appendices~\ref{sec:appa.rq1} and \ref{sec:appa.rq2}.

\begin{figure}[t]
  \centering
  \includegraphics[width=1.0\textwidth]{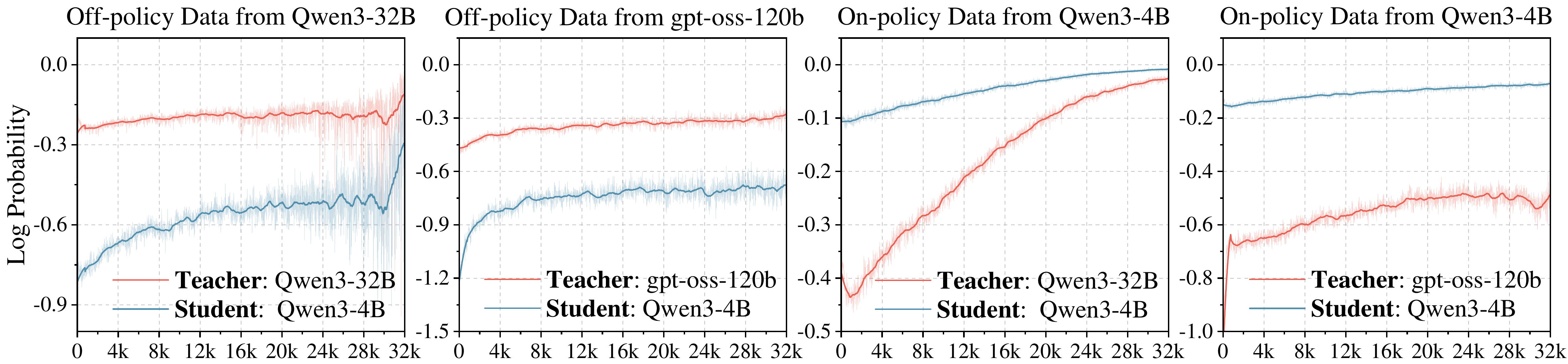}
  \vspace{-8pt}
  \caption{Next-token log probabilities of teacher and student under off-policy and on-policy contexts.}
  \label{logits}
\end{figure}

\begin{figure}[t]
  \centering
  \includegraphics[width=1.0\textwidth]{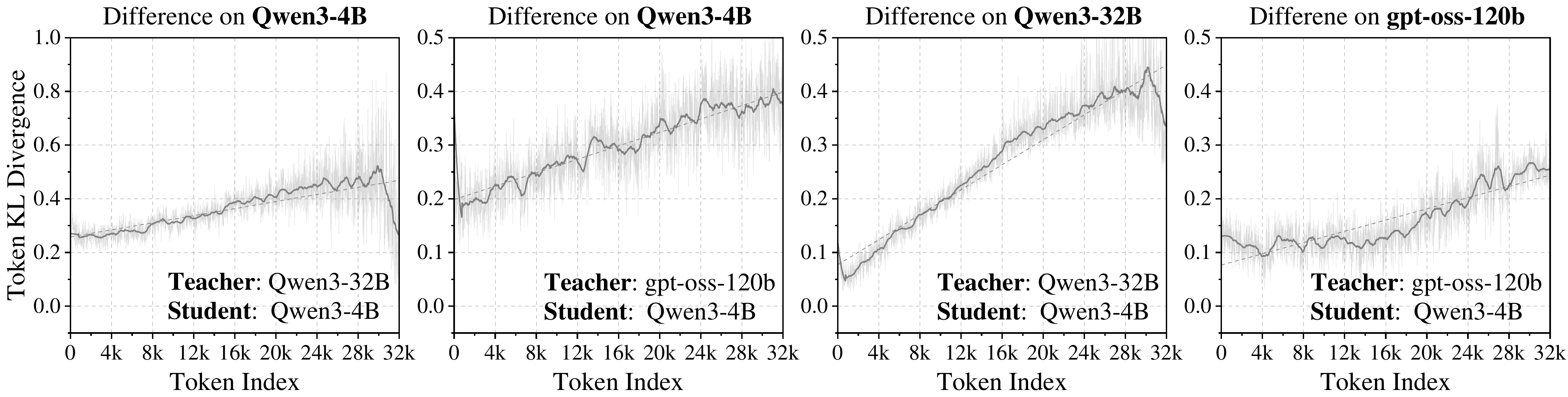}
  \vspace{-8pt}
  \caption{KL divergence between off-policy and on-policy trajectories across context lengths.}
  \label{confidence_difference}
\end{figure}

\subsection{Empirical Analysis of Dual Exposure Biases}
\label{sec:empirical_analysis}

To substantiate the existence of the dual exposure biases, we conduct preliminary experiments using Qwen3-32B and gpt-oss-120b as teacher models, and Qwen3-4B-Instruct, which is initialized via an off-policy distillation phase on datasets corresponding to their respective teachers, as the student.

We first evaluate the models on off-policy data, where the context comprises golden trajectories generated exclusively by the teacher. As illustrated in Fig.~\ref{logits}, the teacher's predictive confidence is consistently high, given that the context perfectly aligns with its native distribution. Conversely, the student model struggles to replicate these exact reasoning steps and exhibits significantly lower confidence.
Furthermore, as shown in Fig.~\ref{confidence_difference}, we illustrate the KL divergence of the next-token probability distributions produced by the student model when conditioned on its own auto-regressively generated on-policy contexts versus teacher-generated off-policy contexts. Generally, this divergence consistently increases as the reasoning sequence lengthens. Therefore, these results clearly illustrate the exposure bias: training solely on golden contexts leaves the student increasingly vulnerable to its own stylistic deviations, causing errors to accumulate rapidly during inference-time generation.

To mitigate exposure bias, on-policy methods rely on the teacher to supervise student-generated trajectories. However, our experiments also uncover a critical \textit{reversed exposure bias}. 
As shown in Fig.~\ref{logits}, when conditioning on sub-optimal on-policy trajectories generated by the student, the student model's step-wise confidence for its self-generated token is consistently higher than the teacher's confidence for the same token. 
Consequently, the guidance provided by the teacher model, based on token probabilities \citep{agarwal2024on}, consistently suppresses these student-generated tokens, thereby hindering the provision of effective positive supervision.
Similarly, in Fig.~\ref{confidence_difference}, the KL divergence of the next-token probability distributions estimated by the teacher model across on-policy and off-policy contexts increases with sequence length, thereby also exhibiting an error accumulation phenomenon.

\subsection{What is the Ideal Unbiased State for Reasoning Distillation?} \label{sec:ideal}

Given the dual exposure biases formulated in Eqs.~\eqref{eq1} and \eqref{eq2} and empirically demonstrated above, an \textit{ideal LLM reasoning distillation process} must satisfy the following two theoretical conditions:
\begin{itemize}
    \item \textbf{\textit{Coverage} for exposure bias}: The training data must be sampled from the student’s intrinsic context distribution $d^{\boldsymbol{\pi}_S}(\mathbf{s})$, encompassing the sub-optimal prefixes the student is likely to encounter during real-world inference to eliminate distribution shift.
    \item \textbf{\textit{Validity} for reversed exposure bias}: The teacher model should be conditioned on a valid context $\mathbf{s}^*$, enabling it to provide informative positive supervision signals.
\end{itemize}
Formally, building upon these two conditions, the ideal reasoning distillation objective aims to minimize the expected negative log-likelihood over this optimal configuration:
\begin{equation}
    \mathcal{J}_\text{ideal}(\boldsymbol{\pi}_S) = \mathbb{E}_{\mathbf{s} \sim d^{\boldsymbol{\pi}_S}} \left[ \mathbb{E}_{\mathbf{y}^* \sim \boldsymbol{\pi}_T(\cdot \mid \mathbf{s}^*)} \left[ -\log \boldsymbol{\pi}_S(\mathbf{y}^* \mid \mathbf{s}) \right] \right].
    \label{eq:ideal_sft}
\end{equation}
In practice, however, this remains challenging, as they exists an inherent dilemma, \ie $\mathbf{s}^* \neq \mathbf{s} \sim d^{\boldsymbol{\pi}_S}$.

\subsection{Can Speculative KD Resolve Dual Exposure Biases?} \label{sec:gap}

In the community, to balance off-policy and on-policy data, some recent works propose \textit{speculative knowledge distillation} (SKD) \citep{xu2025speculative,kim2025in}, where the teacher continuously monitors and corrects the student's on-policy generation. The core intuition is that if an erroneous token $\hat{y}_i$ is produced by the student, it is immediately replaced by a teacher-generated token sampled from $\boldsymbol{\pi}_T(\cdot \mid \mathbf{s}^S)$. The training objective for this pipeline is as follows:
\begin{equation}
    \mathcal{J}_\text{skd}(\boldsymbol{\pi}_S) = \mathbb{E}_{\mathbf{s} \sim d_\text{mix}} \left[ \mathbb{E}_{\hat{y} \sim \boldsymbol{\pi}_T(\cdot \mid \mathbf{s})} [-\log \boldsymbol{\pi}_S(\hat{y} \mid \mathbf{s})] \right],
    \label{eq:direct_sft}
\end{equation}
where $d_\text{mix} = \beta d^{\boldsymbol{\pi}_S} + (1-\beta)d^{\boldsymbol{\pi}_T}$ denotes a mixture distribution between the teacher's and the student's context distributions. 
However, by comparing $\mathcal{J}_\text{skd}$ in Eq.~\eqref{eq:direct_sft} with the ideal objective $\mathcal{J}_\text{ideal}$ in Eq.~\eqref{eq:ideal_sft}, we identify two critical gaps in SKD:
\begin{itemize}
    \item \textbf{Coverage gap} ($d_\text{mix} \neq d^{\boldsymbol{\pi}_S}$): By monitoring the student's generation process step-by-step, SKD strictly corrects the steps where the student deviates from the teacher's distribution, thereby failing to expose the student's sub-optimal trajectories.
    \item \textbf{Validity gap} ($\boldsymbol{\pi}_T(\cdot \mid \mathbf{s}) \neq \boldsymbol{\pi}_T(\cdot \mid \mathbf{s}^*)$): Forced token replacements break the natural auto-regressive coherence of the trajectory, so the teacher $\boldsymbol{\pi}_T$ still suffers from out-of-distribution collapse. Lacking the global error-correction, it will produce high-entropy noise rather than meaningful guidance, violating the validity condition.
\end{itemize}

More formal demonstrations of the failure of this mixture distribution are provided in Appendix~\ref{sec:appa.rq3}. Therefore, to bridge these two gaps and more accurately approximate the ideal objective $\mathcal{J}_\text{ideal}$, we propose a more robust data synthesis framework \baby in this work.

\section{The Proposed \baby}
\label{sec:method}

In this section, we introduce our proposed \baby method that mitigate dual exposure biases.

\noindent
\textbf{Method overview.}
To address the dilemma of the dual exposure biases, the ideal conditions should satisfy both coverage of the student's sub-optimal distribution and validity, ensuring the teacher provides guidance within a relatively correct context. To achieve this, the basic idea of \baby involves consistently monitoring the on-policy trajectories generated by the student, \textit{tolerating sub-optimal paths within an adaptive safety boundary, and backtracking to a relatively safe state for teacher intervention once that boundary is exceeded}. 

\begin{wrapfigure}{r}{0.56\textwidth}
\vspace{-12pt}
    \centering
    \includegraphics[width=0.56\textwidth]{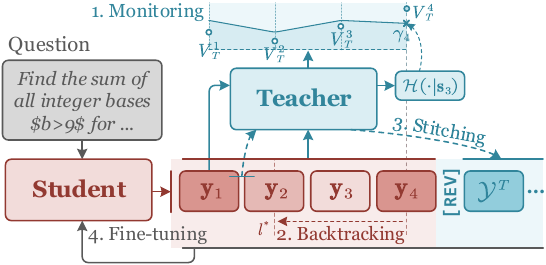}
    \vspace{-10pt}
    \caption{Overall framework of the proposed \baby.}
    \label{framework}
    \vspace{-7pt}
\end{wrapfigure}
Specifically, \baby is an iterative data synthesis framework comprised of three critical steps: 
(1) \textit{\textbf{on-policy trajectory monitoring}}: As the student generates each reasoning step, its likelihood is consistently monitored by the teacher model. Specifically, the entropy of the step generated by the teacher within the given context serves as the adaptive safety boundary. Higher entropy signifies the existence of multiple viable paths toward the correct solution \citep{wang2025beyond}, thereby expanding the safety margin. As long as the student’s trajectory remains within this boundary, its sub-optimal trajectories are tolerated;
(2) \textit{\textbf{State backtracking}}: Once the student violates this safety boundary, a credit assignment mechanism \citep{schulman2016high,xi2026agentprm} is employed to trace back along the on-policy trajectory. This process identifies the specific step characterized by a precipitous decline in likelihood scores, facilitating a backtrack to the last safe state preceding the error;
(3) \textit{\textbf{off-policy trajectory stitching}}: Within this relatively correct context following the backtrack, the teacher model performs a continuation to generate a corrected sequence. Instead of directly using this teacher-generated trajectory as training data, we append the teacher's suffix to the on-policy trajectory precisely at the point where the safety boundary is violated, therefore to explicitly teach the student model how to self-correct its previous deviations.
For clarity, the overall framework of \baby is depicted in Fig.~\ref{framework}. In the following subsections, we formally describe the details of these three steps, respectively.

\subsection{On-policy Trajectory Monitoring via Adaptive Boundary}
\label{subsec:monitoring}

To satisfy the \textit{coverage} condition, the student model must be allowed to explore its intrinsic distribution $d^{\boldsymbol{\pi}_S}$. However, while unconstrained exploration inevitably leads to severe logical hallucinations, excessively restrictive student exploration, \eg SKD \citep{xu2025speculative,yang2026making}, fails to expose the student's sub-optimal distribution. Therefore, \baby monitors the student's step-wise generation using a teacher-guided \textit{adaptive safety boundary}.
Given a reasoning context $\mathbf{s}_{l-1}$ at step $l$, the student policy $\boldsymbol{\pi}_S$ generates a candidate reasoning step $\mathbf{y}_l^S$. To evaluate the validity of this step without aggressively truncating acceptable sub-optimal logic, we define a step-level value function $V_T \in [0,1]$ using the teacher's length-normalized conditional likelihood \citep{xu2025speculative,liu2025where}:
\begin{equation}
    V_T \left( \mathbf{s}_{l-1}, \mathbf{y}_l^S \right) = \exp \left( \frac{1}{|\mathbf{y}_l^S|} \sum \nolimits_{i=1}^{|\mathbf{y}_l^S|} \log \boldsymbol{\pi}_T \left( y_{l,i}^S \mid \mathbf{s}_{l-1}, y_{l,<i}^S \right) \right),
    \label{eq:value_function}
\end{equation}
where $y_{l,i}^S$ denotes the $i$-th token of the reasoning step $\mathbf{y}_l^S$.
Instead of utilizing a rigid, static threshold which artificially restricts the student's exploration, we introduce a state-adaptive boundary $\gamma(\mathbf{s}_{l-1})$. This boundary is inversely proportional to the teacher model's predictive sequence entropy $\mathcal{H}(\boldsymbol{\pi}_T \mid \mathbf{s}_{l-1})$. A high entropy indicates a reasoning manifold with multiple acceptable paths \citep{wang2025beyond,cheng2026reasoning}, warranting a lower threshold, which allows the student to explore more freely; conversely, deterministic axiomatic steps require strict adherence. The adaptive boundary is formulated as
\begin{equation}
    \gamma(\mathbf{s}_{l-1}) = \gamma_0 \cdot \exp \left(-\alpha \cdot \mathcal{H}(\boldsymbol{\pi}_T \mid \mathbf{s}_{l-1}) \right),
    \label{eq:adaptive_gamma}
\end{equation}
where $\gamma_0 \in (0, 1]$ is the base threshold and $\alpha > 0$ is a scaling factor. If $V_T(\mathbf{s}_{l-1}, \mathbf{y}_l^S) \ge \gamma(\mathbf{s}_{l-1})$, the step is deemed safe, and the state transitions normally to $\mathbf{s}_l = [\mathbf{s}_{l-1}, \mathbf{y}_l^S]$. This mechanism guarantees that the student's logically sound, albeit stylistic, explorations are preserved.

\subsection{State Backtracking via Credit Assignment}
\label{subsec:backtracking}

When the student's generation breaches the safety boundary, \ie $V_T < \gamma(\mathbf{s}_{l-1})$, a critical logical collapse is detected. However, directly replacing the immediate step $\mathbf{y}_l^S$ is myopic, as logical errors are often delayed, \eg a flawed premise established at step $l-2$ only manifests as a mathematical contradiction at step $l$. Furthermore, forcing the teacher to correct from the flawed state $\mathbf{s}_{l-1}$ violates the \textit{validity} condition, triggering reversed exposure bias.
To address this, \baby employs a \textit{credit assignment} strategy \citep{lowe2017multi,pignatelli2024a} to backtrack to the safe bifurcation step. Specifically, we compute the step-wise \textit{temporal difference error} $\delta_k$ \citep{schulman2016high,xi2026agentprm} along the historical trajectory to identify the precise moment the logic deviated from the teacher's safe manifold:
\begin{equation}
    \delta_k = V_T \left( \mathbf{s}_{k-1}, \mathbf{y}_k^S \right) - V_T \left( \mathbf{s}_{k-2}, \mathbf{y}_{k-1}^S \right), \quad \forall k \in \{1, \dots, l\}.
\end{equation}
The safe step $l^*$ is identified as the node exhibiting the most severe degradation in oracle value, strictly constrained such that the preceding state remains safe:
\begin{equation}
    l^* = \arg\min_{k \le l} (\delta_k) \quad \text{s.t.} \quad V_T \left( \mathbf{s}_{l^*-2}, \mathbf{y}_{l^*-1}^S \right) \ge \gamma(\mathbf{s}_{l^*-2}).
\end{equation}
By isolating $l^*$, \baby rewinds the reasoning process strictly to the last pristine context $\mathbf{s}_{l^*-1}$. This structurally isolates the teacher from any out-of-distribution sub-optimal steps, ensuring that the subsequent teacher supervision remains uncorrupted by reversed exposure bias.

\subsection{Off-policy Trajectory Stitching}
\label{subsec:stitching}

Having identified the safe prefix $\mathbf{s}_{l^*-1}$, the teacher policy $\boldsymbol{\pi}_T$ is queried to generate a valid, off-policy corrective suffix $\mathcal{Y}_{l^*:T}^T$ as
\begin{equation}
    \mathcal{Y}_{l^*:T}^T \sim \boldsymbol{\pi}_T(\cdot \mid \mathbf{s}_{l^*-1}).
\end{equation}
Because $\mathbf{s}_{l^*-1}$ belongs to the valid support manifold of the teacher, $\boldsymbol{\pi}_T$ provides highly confident and accurate guidance. 
To construct the final trajectory for fine-tuning, instead of directly use $[\mathbf{s}_{l^*-1}, \mathcal{Y}_{l^*:T}^T]$, we seamlessly stitch the student's sub-optimal exploration, an explicit semantic transition, and the teacher's correction. Specifically, we retain the student's fully flawed trajectory up to unsafe step $l$ to act as negative exposure. A discrete semantic revising token, denoted as \texttt{[REV]}, \eg \textit{``However''}, is appended to asynchronously stitch the pristine correction to the flawed context:
\begin{equation}
    \boldsymbol{\tau}_\text{SFT} = \left[ \mathbf{q}, \mathcal{Y}_{1:l}^S \right] \oplus \texttt{[REV]} \oplus \mathcal{Y}_{l^*:T}^T.
    \label{eq:stitching}
\end{equation}
During fine-tuning, by conditioning the student's loss on the sub-optimal on-policy state to predict \texttt{[REV]} and then output the correct steps from $l^*$, the student explicitly learns to: (1) operate robustly within its own generation distribution (mitigating exposure bias), (2) recognize its own logical drift, and (3) perform self-correction using valid oracle targets (bypassing reversed exposure bias).

\subsection{Theoretical Alignment to The Ideal Objective}
\label{subsec:theoretical_alignment}

To theoretically validate \baby, we demonstrate its alignment with the ideal objective $\mathcal{J}_\text{ideal}$ in Eq.~\eqref{eq:ideal_sft} and illustrate how it resolves the \textit{coverage} and \textit{validity} gaps inherent in SKD in Eq.~\eqref{eq:direct_sft}.

In \baby, the student policy is optimized via maximum likelihood estimation over the dynamically stitched trajectories. The expected empirical loss $\mathcal{J}_{\text{\baby}}(\boldsymbol{\pi}_S)$ can be decomposed into a piecewise formulation conditionally dependent on the adaptive boundary:
\begin{equation}
    \begin{aligned}
        \mathcal{J}_{\text{\baby}}(\boldsymbol{\pi}_S) = & \mathbb{E}_{\mathbf{s}_{l-1} \sim d^{\boldsymbol{\pi}_S}} \mathbb{E}_{\mathbf{y}_l^S \sim \boldsymbol{\pi}_S} 
        \Big[ \mathbb{I}_{[V_T \ge \gamma(\mathbf{s})]} \left[-\log \boldsymbol{\pi}_S \left( \mathbf{y}_l^S \mid \mathbf{s}_{l-1} \right) \right] \\
        & + \mathbb{I}_{[V_T < \gamma(\mathbf{s})]} \mathbb{E}_{\mathcal{Y}^T \sim \boldsymbol{\pi}_T(\cdot \mid \mathbf{s}_{l^*-1})} 
        \left[ -\log \boldsymbol{\pi}_S \left( \texttt{[REV]} \oplus \mathcal{Y}^T \mid \mathbf{s}_{l-1} \oplus \mathbf{y}_t^S \right) \right] \Big].
    \end{aligned}
    \label{eq:baby_objective}
\end{equation}
By comparing Eq.~\eqref{eq:baby_objective} with $\mathcal{J}_\text{ideal}$, we show that \baby strictly satisfies both prerequisite conditions:
Unlike SKD which restricts states to a narrow mixture $d_\text{mix}$, Eq.~\eqref{eq:baby_objective} operates strictly over $\mathbf{s}_{l-1} \sim d^{\boldsymbol{\pi}_S}$ and $\mathbf{y}_t^S \sim \boldsymbol{\pi}_S$. Retaining the flawed context $(\mathbf{s}_{l-1} \oplus \mathbf{y}_t^S)$ as the prefix ensures full coverage of the student's authentic inference manifold, effectively eliminating the distribution shift.
Forcing the teacher to predict upon a flawed context $\mathbf{s}_t$ induces out-of-distribution noise ($\boldsymbol{\pi}_T \neq \boldsymbol{\pi}_T^*$). \baby approximates the ideal oracle $\boldsymbol{\pi}_T^*$ by sampling the correction $\mathcal{Y}^T$ exclusively from the pristine state $\mathbf{s}_{l^*-1} \in \text{supp}(d^{\boldsymbol{\pi}_T})$. Stitching via the \texttt{[REV]} token guarantees the teacher's supervision remains valid and highly confident.
We provide a more rigorous theoretical analysis of \baby in Appendix~\ref{sec:appa.rq4}.

Accordingly, $\mathcal{J}_{\text{\baby}}$ serves as a robust estimator for $\mathcal{J}_\text{ideal}$, simultaneously mitigating the dual exposure biases through structurally sound data synthesis.

\vspace{-3pt}
\section{Experimental Evaluation} \label{sec:experiments}
\vspace{-3pt}

\noindent
\textbf{Experimental settings.}
Our problem sets are derived from the \textit{LIMO-v2} repository \citep{ye2025limo} and a randomly sampled subset of 10,000 questions from the \textit{AceReason} repository \citep{liu2025acereason}. We evaluate our trained student models across three mathematical reasoning benchmarks: AIME24, AIME25, and MATH500 \citep{lightman2024let}, and two cross-domain benchmarks: IFEval \citep{zhou2023instruction} and GPQA \citep{rein2023gpqa}. For AIME24, AIME25, and GPQA, we sample 8 times per problem and report the pass@8, pass@4, and pass@1 metrics.
We utilize Qwen3-32B and gpt-oss-120b as teacher models, while Qwen3-4B-Instruct, Qwen2.5-7B-Instruct, and Llama3.2-3B serve as the student models. All experiments were conducted on an 8-GPU A800 cluster, employing the SGLang framework for data generation and LlamaFactory for model training. 
Our selected baselines include \textit{supervised fine-tuning} (SFT), \textit{rejection sampling fine-tuning} (RFT), ImitKD \citep{lin2020autoregressive}, and SKD \citep{xu2025speculative}. Specifically, ImitKD involves randomly selecting steps from either the student or the teacher during the decoding process. In contrast, SKD accepts a student-generated step only if the log probability assigned by the teacher exceeds the student's log probability by a factor of $\beta$; otherwise, the step is rejected, and the teacher generates a replacement.
More detailed settings, \eg LLM model cards and implementation details, can be seen in Appendix~\ref{app.c}.

\setlength{\aboverulesep}{0.2ex}
\setlength{\belowrulesep}{0.2ex}
\begin{table}[t]
\centering
\renewcommand{\arraystretch}{1.10}
\setlength{\tabcolsep}{4pt}
\caption{Experimental results using Qwen3-4B-Instruct as the student model.}
\label{mainresults}
\small
    \begin{tabular}{c m{1.90cm}<{\centering}m{0.66cm}<{\centering}m{0.66cm}<{\centering}m{0.66cm}<{\centering}m{0.66cm}<{\centering}m{0.66cm}<{\centering}m{0.66cm}<{\centering}m{0.66cm}<{\centering}m{0.66cm}<{\centering}m{0.66cm}<{\centering}m{0.66cm}<{\centering}m{0.66cm}<{\centering}m{0.75cm}<{\centering}}
    \toprule
   \multirow{2}{*}{\footnotesize \rotatebox{90}{Teacher}} & \multirow{2}{*}{Method} & \multicolumn{3}{c}{AIME24} & \multicolumn{3}{c}{AIME25} & \multicolumn{5}{c}{MATH500} & \multirow{2}{*}{Avg.} \\
    \cmidrule(r){3-5} \cmidrule(r){6-8} \cmidrule(r){9-13}
    & & P@8 & P@4 & P@1 & P@8 & P@4 & P@1 & L1 & L2 & L3 & L4 & L5 &  \\
    \midrule
    \multirow{7}{*}{\rotatebox{90}{\textbf{Qwen3-32B}}}
    & SFT & 50.00 & 37.81 & 21.25 & 40.00 & 35.43 & 24.17 & 90.70 & 82.22 & 77.14 & 70.31 & 51.49 & 52.77 \\
    & RFT & 46.67 & 38.38 & 21.67 & 40.00 & 30.95 & 21.67 & 83.72 & 82.22 & 85.71 & 65.62 & 51.49 & 51.65 \\
    & ImitKD & 46.67 & 40.71 & 24.17 & 40.00 & 33.05 & 20.00 & 93.02 & 82.22 & 87.62 & 77.34 & 60.45 & 55.02 \\
    & SKD {\footnotesize ($\beta=0.6$)} & 53.33 & 45.00 & 25.00 & 46.67 & 43.33 & 25.00 & 88.37 & 85.55 & 89.52 & 75.78 & 60.45 & 58.09 \\
    & SKD {\footnotesize ($\beta=0.8$)} & 53.33 & 48.33 & 24.17 & 46.67 & 45.00 & 27.50 & 88.37 & 86.67 & 87.61 & 82.03 & 63.43 & 59.37 \\
    & SKD {\footnotesize ($\beta=1.0$)} & 56.67 & 43.33 & 27.50 & 46.67 & 40.00 & 25.83 & 90.07 & 86.67 & 88.57 & 81.25 & 56.72 & 58.47 \\
    \cmidrule{2-14}
    & \cellcolor{lightgrayv} \textbf{\baby} {\footnotesize (ours)} & \cellcolor{lightgrayv} \textbf{66.67} & \cellcolor{lightgrayv} \textbf{57.57} & \cellcolor{lightgrayv} \textbf{34.17} & \cellcolor{lightgrayv} \textbf{56.67} & \cellcolor{lightgrayv} \textbf{50.71} & \cellcolor{lightgrayv} \textbf{31.67} & \cellcolor{lightgrayv} \textbf{93.02} & \cellcolor{lightgrayv} \textbf{86.67} & \cellcolor{lightgrayv} \textbf{95.24} & \cellcolor{lightgrayv} \textbf{86.72} & \cellcolor{lightgrayv} \textbf{73.88} & \cellcolor{lightgrayv} \textbf{66.64} \\
    \hline
    \specialrule{0em}{0.5pt}{0.5pt}
    \hline
    \multirow{7}{*}{\rotatebox{90}{\textbf{gpt-oss-120b}}}
    & SFT & 76.67 & 73.14 & 43.33 & 73.33 & 61.48 & 37.50 & 90.70 & 90.00 & 80.00 & 74.22 & 50.75 & 68.28 \\
    & ImitKD & 76.67 & 72.29 & 44.17 & 63.33 & 57.29 & 37.92 & 81.40 & 67.78 & 71.43 & 69.53 & 61.19 & 63.91 \\
    & SKD {\footnotesize ($\beta=0.6$)} & 80.00 & 74.57 & 45.83 & 66.67 & 61.48 & 39.58 & 69.77 & 74.44 & 77.14 & 71.88 & 72.39 & 66.70 \\
    & SKD {\footnotesize ($\beta=0.8$)} & 83.33 & 75.29 & 46.66 & 80.00 & 72.52 & 45.00 & 90.70 & 92.22 & 84.76 & 81.25 & 74.63 & 75.12 \\
    & SKD {\footnotesize ($\beta=1.0$)} & 80.00 & 73.14 & 36.25 & 80.00 & 64.33 & 33.75 & 74.42 & 72.22 & 71.43 & 75.00 & 73.13 & 66.70 \\
    \cmidrule{2-14}
    & \cellcolor{lightgrayv} \textbf{\baby} {\footnotesize (ours)} & \cellcolor{lightgrayv} \textbf{86.67} & \cellcolor{lightgrayv} \textbf{78.33} & \cellcolor{lightgrayv} \textbf{47.08} & \cellcolor{lightgrayv} \textbf{83.33} & \cellcolor{lightgrayv} \textbf{74.14} & \cellcolor{lightgrayv} \textbf{47.50} & \cellcolor{lightgrayv} \textbf{93.02} & \cellcolor{lightgrayv} \textbf{95.56} & \cellcolor{lightgrayv} \textbf{88.57} & \cellcolor{lightgrayv} \textbf{82.81} & \cellcolor{lightgrayv} \textbf{76.87} & \cellcolor{lightgrayv} \textbf{77.62} \\
    \bottomrule
    \end{tabular}
\vspace{-3pt}
\end{table}

\vspace{-3pt}
\subsection{Main Results}
\label{subsec:main_results}
\vspace{-3pt}

Table~\ref{mainresults} summarizes the results of our proposed \baby against various baseline methods. Overall, \baby consistently and significantly outperforms all baselines across different teacher models and metrics. Notably, when utilizing Qwen3-32B as the teacher, \baby achieves an average accuracy of 66.64\%, yielding a substantial improvement of over 7\% compared to the strongest baseline SKD. This superiority is strictly maintained when scaling up to the gpt-oss-120b teacher. Furthermore, our method exhibits remarkable resilience on highly complex tasks requiring long CoT trajectories. On challenging benchmarks, \eg AIME24 and AIME25, especially under the strict Pass@1 metric, and the most difficult subsets of MATH500, standard off-policy methods, \eg SFT and RFT, struggle severely due to error cascades caused by exposure bias. Similarly, aggressive on-policy interventions like ImitKD and SKD yield limited gains, as they force the teacher to generate from sub-optimal contexts, exacerbating reversed exposure bias. In contrast, \baby drastically boosts performance by adaptively tolerating sub-optimal explorations and backtracking to the safe point for intervention.

Beyond the primary evaluations, we extensively validate the robustness and generalizability of our framework across varied data sources, student architectures, and downstream domains. As shown in Table~\ref{acereason}, training on the distinct \textit{AceReason} dataset yields identically superior trends, confirming that the efficacy of \baby is independent of specific data distributions. Moreover, Table~\ref{more_student_1} demonstrates that our method seamlessly transfers to diverse student architectures, consistently elevating the reasoning capabilities of both Qwen2.5-7B and Llama3.2-3B. Finally, to assess out-of-domain transferability, we evaluate the math-trained models on instruction following, \ie IFEval, and scientific reasoning, \ie GPQA. As detailed in Table~\ref{more_student_2}, \baby achieves state-of-the-art cross-domain performance regardless of the teacher model used. This verifies that our trajectory monitoring and backtracking mechanisms not only prevent mathematical error cascades but also instill a generalized, domain-agnostic self-reflection ability that broadly enhances complex, multi-step problem-solving.

Table~\ref{onpolicy} compares the performance of \baby against several on-policy distillation approaches, \eg top-$k$ OPD and DASD. Since on-policy methods are highly memory-intensive and require vocabulary alignment between the teacher and student models, we employ Qwen3-8B as the teacher and Qwen3-1.7B as the student. The results demonstrate that our method not only significantly outperforms off-policy baselines but also remains comparable to on-policy methods. Furthermore, given that on-policy methods entail substantially higher time complexity and GPU memory consumption than our approach, these results further validate the effectiveness and efficiency of our method.

\begin{table}[t]
    \centering
    \begin{minipage}{0.5\textwidth}
        \centering
        \renewcommand{\arraystretch}{1.0}
        \setlength{\tabcolsep}{4pt}
        \caption{Ablation study of \baby.}
        \label{tab:ablation} 
        \footnotesize
        \begin{tabular}{m{1.35cm}<{\centering}m{0.75cm}<{\centering}m{0.75cm}<{\centering}m{0.75cm}<{\centering}m{0.75cm}<{\centering}m{0.75cm}<{\centering}}
            \toprule
            \multirow{2}{*}{Method} & \multicolumn{2}{c}{AIME24} & \multicolumn{2}{c}{AIME25} & \multirow{2}{*}{MATH} \\
            \cmidrule(r){2-3} \cmidrule(r){4-5}
            & P@8 & P@1 & P@8 & P@1 &  \\
            \hline
            \rowcolor{lightgrayv} \baby & \textbf{66.67} & \textbf{34.17} & \textbf{56.67} & \textbf{31.67} & \textbf{85.60} \\
            w/o adapt & 63.33 & 32.50 & 50.00 & 29.16 & 84.00 \\
            w/o back & 56.67 & 30.83 & 50.00 & 27.91 & 82.80 \\
            w/o stitch & 63.33 & 32.08 & 53.33 & 29.58 & 83.60 \\
            \bottomrule
        \end{tabular}
    \end{minipage}%
    \hfill 
    \begin{minipage}{0.5\textwidth}
        \centering
        \renewcommand{\arraystretch}{1.0}
        \setlength{\tabcolsep}{4pt}
        \caption{Results on 10k questions from \textit{AceReason}.}
        \label{acereason} 
        \footnotesize
        \begin{tabular}{m{1.1cm}<{\centering}m{0.75cm}<{\centering}m{0.75cm}<{\centering}m{0.75cm}<{\centering}m{0.75cm}<{\centering}m{0.75cm}<{\centering}}
            \toprule
            \multirow{2}{*}{Method} & \multicolumn{2}{c}{AIME24} & \multicolumn{2}{c}{AIME25} & \multirow{2}{*}{MATH} \\
            \cmidrule(r){2-3} \cmidrule(r){4-5}
            & P@8 & P@1 & P@8 & P@1 &  \\
            \hline
            SFT & 56.67 & 25.41 & 53.33 & 23.33 & 81.00 \\
            ImitKD & 63.33 & 31.25 & 53.33 & 28.33 & 81.20 \\
            SKD & 63.33 & 32.92 & 53.33 & 28.75 & 82.60 \\
            \rowcolor{lightgrayv} \baby & \textbf{70.00} & \textbf{35.00} & \textbf{56.67} & \textbf{30.42} & \textbf{85.80} \\
            \bottomrule
        \end{tabular}
    \end{minipage}
\end{table}

\subsection{Ablation and Sensitivity Analysis}
\label{subsec:ablation}

\noindent
\textbf{Ablation study.}
To validate our framework's core components, we conduct an ablation study on the AIME24, AIME25, and MATH benchmarks. As shown in Table~\ref{tab:ablation}, removing any module consistently degrades performance. Crucially, disabling state backtracking, \ie \textit{w/o back}, incurs the most severe decline, \eg dropping from 34.17\% to 30.83\% on AIME24 Pass@1. This highlights that forcing immediate teacher intervention without tracing the logical root cause pollutes the context, thereby triggering reversed exposure bias. Furthermore, omitting sub-optimal trajectory stitching, \ie \textit{w/o stitch}, via direct sequence truncation noticeably impairs accuracy. This confirms that explicitly exposing the student to its own sub-optimal prefixes is vital for learning self-correction and mitigating exposure bias. Finally, substituting the entropy-based adaptive boundary with a static threshold, \ie \textit{w/o adapt}, also harms performance, demonstrating that rigid constraints cannot effectively balance student exploration diversity with the validity of teacher supervision.

\begin{table}[t]
    \centering
    \begin{minipage}{0.48\textwidth}
        \centering
        \vspace{-10pt}
        \renewcommand{\arraystretch}{1.15}
        \setlength{\tabcolsep}{4pt}
        \caption{Results of on-policy methods.}
        \label{onpolicy}
        \footnotesize
        \begin{tabular}{m{1.75cm}<{\centering}m{0.65cm}<{\centering}m{0.65cm}<{\centering}m{0.65cm}<{\centering}m{0.65cm}<{\centering}m{0.8cm}<{\centering}}
            \toprule
            \multirow{2}{*}{Method} & \multicolumn{2}{c}{AIME24} & \multicolumn{2}{c}{AIME25} & \multirow{2}{*}{MATH} \\
            \cmidrule(r){2-3} \cmidrule(r){4-5}
            & P@8 & P@1 & P@8 & P@1 &  \\
            \hline
            SFT & 50.00 & 18.75 & \textbf{60.00} & 21.25 & 69.00 \\
            RFT & 13.33 & 5.83 & 33.33 & 12.08 & 64.20 \\
            SKD & 53.33 & 20.41 & 56.67 & 22.08 & 71.40 \\
            \rowcolor{lightgrayv} top-$k$ OPD \citep{fu2026revisiting} & \textbf{60.00} & 32.91 & 53.33 & 30.41 & \textbf{79.20} \\
            \rowcolor{lightgrayv} DASD \citep{yan2026distribution} & 53.33 & 26.67 & 53.33 & 28.75 & 74.60 \\
            \hline
            \rowcolor{lightgrayv} \baby & \textbf{60.00} & \textbf{33.75} & \textbf{60.00} & \textbf{31.67} & 79.00 \\
            \bottomrule
        \end{tabular}
    \end{minipage}
    \hfill
    \begin{minipage}{0.48\textwidth}
        \centering
        \includegraphics[width=0.90\linewidth]{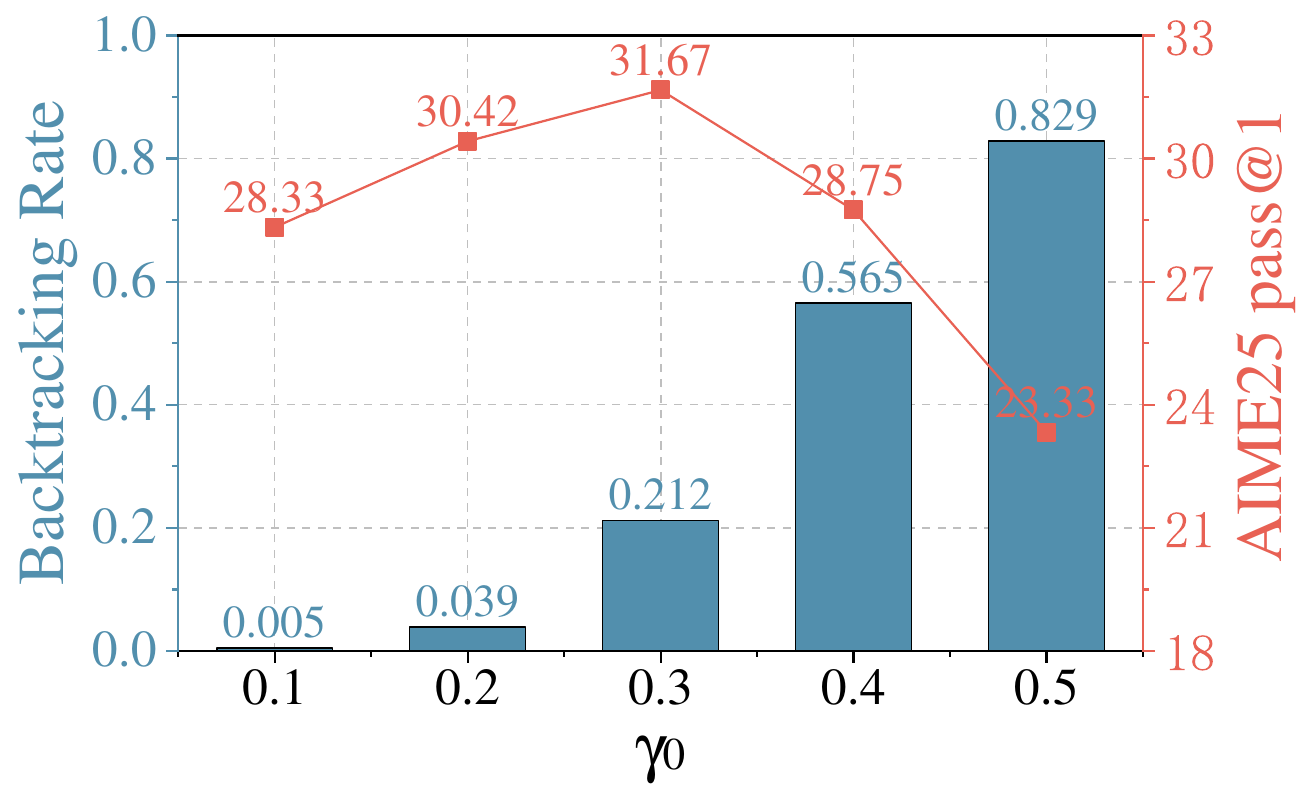} 
        \caption{Sensitivity analysis of the parameter $\gamma_0$.}
        \label{fig:gamma_sensitivity}
    \end{minipage}
\vspace{-5pt}
\end{table}

\noindent
\textbf{Sensitivity analysis.}
We conduct a sensitivity analysis on the threshold $\gamma_0$ to evaluate its impact on reasoning performance and intervention frequency. As illustrated in Fig.~\ref{fig:gamma_sensitivity}, increasing $\gamma_0$ inherently enforces a stricter safety boundary, leading to a monotonic rise in the backtracking rate. However, the reasoning performance on AIME25 Pass@1 exhibits an inverted-U trend, peaking at a moderate value of $\gamma_0 = 0.3$. This suggests that an overly low $\gamma_0$ is too lenient, failing to catch severe logical drifts and thus permitting error cascades. Conversely, an excessively high $\gamma_0$ becomes overly rigid, constantly interrupting the student's natural exploration and discarding acceptable sub-optimal paths, which undermines the mitigation of exposure bias.

\begin{table}[t]
    \centering
    \begin{minipage}{0.48\textwidth}
        \centering
        \renewcommand{\arraystretch}{1.05}
        \setlength{\tabcolsep}{4pt}
        \caption{Performance on more student models.}
        \label{more_student_1} 
        \footnotesize
        \begin{tabular}{c m{1.0cm}<{\centering}m{0.7cm}<{\centering}m{0.7cm}<{\centering}m{0.7cm}<{\centering}m{0.7cm}<{\centering}m{0.7cm}<{\centering}}
            \toprule
            & \multirow{2}{*}{Method} & \multicolumn{2}{c}{AIME24} & \multicolumn{2}{c}{AIME25} & \multirow{2}{*}{MATH} \\
            \cmidrule(r){3-4} \cmidrule(r){5-6}
            & & P@8 & P@1 & P@8 & P@1 &  \\
            \hline
            \multirow{4}{*}{\scriptsize \rotatebox{90}{\textbf{Qwen2.5-7B}}}
            & SFT & 23.33 & \textbf{10.42} & 16.67 & 4.58 & 62.60 \\
            & ImitKD & 20.00 & 5.83 & \textbf{26.67} & 5.83 & 58.60 \\
            & SKD & 23.33 & 8.06 & 24.44 & 7.36 & 64.27 \\
            & \cellcolor{lightgrayv} \baby & \cellcolor{lightgrayv} \textbf{30.00} & \cellcolor{lightgrayv} \textbf{10.42} & \cellcolor{lightgrayv} \textbf{26.67} & \cellcolor{lightgrayv} \textbf{7.92} & \cellcolor{lightgrayv} \textbf{70.40} \\
            \hline
            \multirow{4}{*}{\scriptsize \rotatebox{90}{\textbf{Llama3.2-3B}}}
            & SFT & 16.67 & 3.33 & 6.67 & 0.83 & 44.00 \\
            & ImitKD & 16.67 & 2.92 & 6.67 & 1.67 & 45.80 \\
            & SKD & 17.78 & 2.92 & 8.89 & 2.22 & 46.47 \\
            & \cellcolor{lightgrayv} \baby & \cellcolor{lightgrayv} \textbf{23.33} & \cellcolor{lightgrayv} \textbf{5.00} & \cellcolor{lightgrayv} \textbf{16.67} & \cellcolor{lightgrayv} \textbf{5.83} & \cellcolor{lightgrayv} \textbf{52.60} \\
            \bottomrule
        \end{tabular}
    \end{minipage}%
    \hfill 
    \begin{minipage}{0.48\textwidth}
        \centering
        \renewcommand{\arraystretch}{1.05}
        \setlength{\tabcolsep}{4pt}
        \caption{Results on datasets from other domains.}
        \label{more_student_2} 
        \footnotesize
        \begin{tabular}{c m{1.0cm}<{\centering}m{0.7cm}<{\centering}m{0.7cm}<{\centering}m{0.7cm}<{\centering}m{0.7cm}<{\centering}m{0.7cm}<{\centering}}
            \toprule
            & \multirow{2}{*}{Method} & \multicolumn{2}{c}{IFEval} & \multicolumn{3}{c}{GPQA} \\
            \cmidrule(r){3-4} \cmidrule(r){5-7}
            & & P.S. & Avg. & P@8 & P@4 & P@1 \\
            \hline
            \multirow{4}{*}{\scriptsize \rotatebox{90}{\textbf{Qwen3-32B}}}
            & SFT & 29.92 & 37.23 & 85.35 & 68.35 & 31.06 \\
            & ImitKD & 32.53 & 41.02 & 84.85 & 68.34 & 39.93 \\
            & SKD & 35.86 & 43.38 & 85.86 & 73.97 & 41.98 \\
            & \cellcolor{lightgrayv} \baby & \cellcolor{lightgrayv} \textbf{37.89} & \cellcolor{lightgrayv} \textbf{46.80} & \cellcolor{lightgrayv} \textbf{89.39} & \cellcolor{lightgrayv} \textbf{77.18} & \cellcolor{lightgrayv} \textbf{46.97} \\
            \hline
            \multirow{4}{*}{\scriptsize \rotatebox{90}{\textbf{gpt-oss-120b}}}
            & SFT & 32.16 & 40.37 & 84.34 & 72.60 & 44.44 \\
            & ImitKD & 32.35 & 40.15 & 76.26 & 65.12 & 39.39 \\
            & SKD & 34.57 & 42.71 & 88.89 & 79.14 & 44.38 \\
            & \cellcolor{lightgrayv} \baby & \cellcolor{lightgrayv} \textbf{41.59} & \cellcolor{lightgrayv} \textbf{50.01} & \cellcolor{lightgrayv} \textbf{91.41} & \cellcolor{lightgrayv} \textbf{79.42} & \cellcolor{lightgrayv} \textbf{50.51} \\
            \bottomrule
        \end{tabular}
    \end{minipage}
\vspace{-7pt}
\end{table}


\vspace{-3pt}
\subsection{Evaluating Dual Exposure Biases}
\vspace{-3pt}

\begin{wrapfigure}{r}{0.5\textwidth}
\vspace{-5pt}
    \centering
    \includegraphics[width=0.505\textwidth]{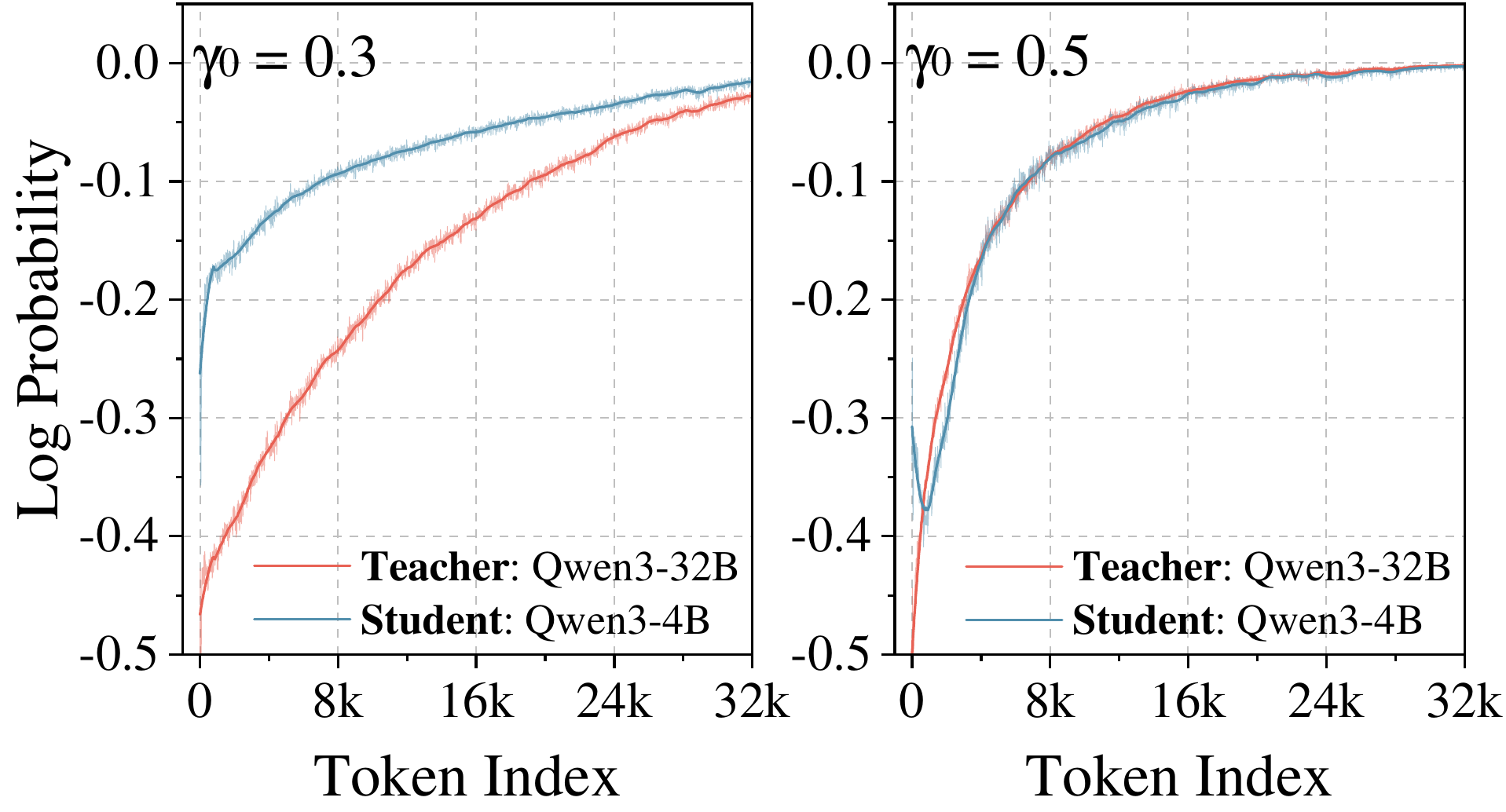}
    \vspace{-10pt}
    \caption{Evaluation of dual exposure biases.}
    \label{fig:dual_bias}
    \vspace{-5pt}
\end{wrapfigure}

To explicitly visualize the mitigation of dual exposure biases, we track the token-level log probabilities of both the student and teacher models along the trajectories synthesized by \baby. As illustrated in Fig.~\ref{fig:dual_bias}, our state backtracking mechanism ensures that the teacher's confidence remains stable, successfully averting the severe probability collapse typically associated with reversed exposure bias. 

Under a boundary $\gamma_0=0.3$, a distributional gap persists, preserving the student's exploratory diversity to mitigate exposure bias while maintaining valid teacher supervision.
Remarkably, when the boundary is tightened to $\gamma_0=0.5$, the probability distributions of the student and teacher align almost perfectly. This empirical evidence demonstrates that \baby effectively harmonizes the student's authentic generation manifold with the teacher's valid supervision manifold, fundamentally resolving the distributional discrepancies in dual exposure biases.

\vspace{-4pt}
\section{Related Works}
\vspace{-4pt}

In this section, we briefly summarize the literature on LLM reasoning distillation.
Generally, prior reasoning distillation methods follow an \textit{off-policy paradigm} \citep{ho2023large,hsieh2023distilling,magister2023teaching,kou2026positive}. This process involves sampling complex problems \citep{ye2025limo,muennighoff2025s1}, utilizing a teacher model to generate multiple responses via temperature sampling \citep{chen2023mcc,lei2025learning,yan2026distribution}, and conducting rigorous filtering based on criteria, \eg response length \citep{wu2025beyond,liu2025acereason}, answer correctness \citep{chen2025skip,li2025naturalthoughts}, and the student model’s adaptability \citep{yuan2024instance,zhang2025dylan,wang2026on}. 

However, as analyzed in Sec.~\ref{sec.2}, early research also identified that such off-policy methods suffer from exposure bias, \ie train-test mismatch \citep{ranzato2016sequence,lin2020autoregressive}.
Theoretically motivated by imitation learning \citep{ross2011a,zhang2017query}, recent methods propose \textit{on-policy distillation} \citep{agarwal2024on,song2026a}. Unlike previous approaches, this paradigm allows the student model to generate its own reasoning trajectories while receiving dense, real-time feedback from the teacher model at each token level, \eg $f$-divergence \citep{agarwal2024on}, log probabilities \citep{gu2024minillm}, and the combination of forward and backward KL divergences \citep{jung2025todi}. In parallel, recent reinforcement learning methods have further demonstrated the efficacy of on-policy data through rollouts by LLM itself \citep{shao2024deepseekmath,chen2025retaining}. In this work, we address the dilemma between off-policy and on-policy distillation, \ie \textit{dual exposure biases}. The most intuitive approach involves identifying a mixture distribution of the two, \eg through the scheduled sampling \citep{bengio2015scheduled}. Recent speculative distillation strategies \citep{xu2025speculative,kim2025in} have similarly adopted this rationale; however, our analysis in Sec.~\ref{sec:gap} demonstrates that such schemes still fail to satisfy the ideal unbiased condition, which ultimately motivates our proposed \baby.

\vspace{-4pt}
\section{Conclusion} \label{sec:conclusion}
\vspace{-4pt}

In this paper, we identify a dilemma between the off-policy and on-policy paradigms in LLM reasoning distillation, \ie \textit{dual exposure biases}. We demonstrate that while off-policy distillation suffers from exposure bias due to training-inference mismatches, on-policy distillation inevitably introduces a reversed exposure bias, wherein the teacher model fails to provide reliable supervision when conditioned on straying, student-generated trajectories. To resolve this dilemma, we introduced \baby, a novel trajectory monitoring and backtracking framework. By employing an adaptive safety boundary, \baby allows the student model to explore and learn from its own errors, thereby mitigating exposure bias, while explicitly backtracking and applying teacher intervention before the context degrades irreparably, circumventing reversed exposure bias. Extensive experiments demonstrate that \baby enhances reasoning capabilities and mitigates dual exposure biases.

\section*{Acknowledgements}

This work was supported by the National Natural Science Foundation of China (No.62276113) and Alibaba Research Intern Program.

\bibliography{reference}
\bibliographystyle{abbrvnat}

\newpage
\appendix

\section{Theoretical Analysis} \label{app:theory_detailed}

In this section, we provide a formal and mathematical derivation of the dual exposure biases in LLM reasoning distillation, and prove how our proposed \baby framework theoretically resolves them. 
Specifically, we answer the following three basic questions:

\newcommand{\hangingpar}[1]{%
  \noindent
  \hangafter=1
  \setlength{\hangindent}{2.5em}
  #1\par
}
\hangingpar{
\quad \textcolor{Blue}{\textbf{RQ1.}} Why does off-policy distillation fail during inference?
}
\hangingpar{
\quad \textcolor{Blue}{\textbf{RQ2.}} Why does on-policy distillation corrupt the teacher's supervision?
}
\hangingpar{
\quad \textcolor{Blue}{\textbf{RQ3.}} Why does a mixture of off-policy and on-policy distributions fail to resolve this dilemma?
}
\hangingpar{
\quad \textcolor{Blue}{\textbf{RQ4.}} How does \baby formally guarantee distribution coverage and supervision validity?
}
LLM reasoning is a strict full-history-dependent auto-regressive process.
Formally, let $\mathcal{V}$ denote the vocabulary space. For an initial question $\mathbf{q}$, a reasoning trajectory of length $L$ is a sequence of reasoning steps $\mathcal{Y} = (\mathbf{y}_1, \mathbf{y}_2, \dots, \mathbf{y}_L)$, and each step is a sequence of tokens $\mathbf{y}_l = (y_{l,1}, y_{l,1}, \ldots, y_{l,|\mathbf{y}_l|})$. 
At any step $l$, the context is the entire historical prefix $\mathbf{s}_{t-1} = [\mathbf{q}, \mathbf{y}_1, \mathbf{y}_2, \ldots, \mathbf{y}_{l-1}]$. 
A policy LLM $\boldsymbol{\pi}$ defines a conditional probability distribution over the next token $\boldsymbol{\pi}(y_{l,i} \mid \mathbf{s}_{t-1}, \mathbf{y}_{l,<i})$. 

The probability of a reasoning trajectory under policy $\boldsymbol{\pi}$ is given by the chain rule of probability:
\begin{equation}
    P_{\boldsymbol{\pi}}(\mathcal{Y} \mid \mathbf{q}) = \prod \nolimits_{l=1}^L \prod \nolimits_{i=1}^{|\mathbf{y}_l|} \boldsymbol{\pi} (y_{l,i} \mid \mathbf{s}_{l-1},\mathbf{y}_{l,<i}). \nonumber
\end{equation}
Let $d^{\boldsymbol{\pi}}$ denote the marginal distribution of all possible historical prefixes $\mathbf{s}$ generated by iteratively unrolling policy $\boldsymbol{\pi}$. 

\subsection{RQ1: Exposure Bias in Off-Policy Distillation} \label{sec:appa.rq1}

Off-policy distillation minimizes the single-step Kullback-Leibler divergence exclusively over the teacher's historical prefixes. We assume the student model $\boldsymbol{\pi}_S$ is optimized such that its expected single-step error on the teacher's manifold $p_T$ is bounded by $\epsilon$:
\begin{equation}
    \mathbb{E}_{\mathbf{s}_{l-1} \sim p_T} \left[ D_{\mathrm{TV}} \big( \boldsymbol{\pi}_T(\cdot \mid \mathbf{s}_{l-1}) \parallel \boldsymbol{\pi}_S(\cdot \mid \mathbf{s}_{l-1}) \big) \right] \le \epsilon, \quad \forall l \in [1, L].
    \label{eq:off_policy_bound}
\end{equation}

\textbf{Theorem 1.} \textit{Under the training condition in Eq.~\eqref{eq:off_policy_bound}, during free-form inference where prefixes are generated by the student $p_S$, the sequence-level divergence $D_{\mathrm{TV}}\big(p_T(\mathbf{s}_T) \parallel p_S(\mathbf{s}_T)\big)$ scales at an $\mathcal{O}(L^2)$ rate in the worst case, demonstrating the exposure bias.}

\noindent
\textbf{Proof.}
Using the chain rule of probability, the prefix distributions can be recursively decomposed: $p(\mathbf{s}_l) = p(\mathbf{s}_{l-1}) \boldsymbol{\pi}(\mathbf{y}_l \mid \mathbf{s}_{l-1})$, where $\boldsymbol{\pi}(\mathbf{y}_l \mid \mathbf{s}_{l-1}) = \prod_{i=1}^{|\mathbf{y}_l|} \boldsymbol{\pi}(y_{l,i} \mid \mathbf{s}_{l-1}, \mathbf{y}_{l,<i})$. 
By the telescoping property of the total variation distance for sequence distributions, the trajectory-level divergence at the final step $L$ can be expanded as the sum of step-level conditional divergences evaluated over the \textit{student's} prefix distribution:
\begin{equation}
    D_{\mathrm{TV}} \big( p_T(\mathbf{s}_L) \parallel p_S(\mathbf{s}_L) \big) 
    \le \sum \nolimits_{l=1}^L \mathbb{E}_{\mathbf{s}_{l-1} \sim p_S} \left[ D_{\mathrm{TV}} \big( \boldsymbol{\pi}_T(\cdot \mid \mathbf{s}_{l-1}) \parallel \boldsymbol{\pi}_S(\cdot \mid \mathbf{s}_{l-1}) \big) \right] \\
    = \sum \nolimits_{l=1}^L \mathcal{L}_{\mathrm{FR}}^{(l)}.
\label{eq:telescoping_tv}
\end{equation}
Here, $\mathcal{L}_{\mathrm{FR}}^{(l)}$ denotes the free-running expected divergence. Notice that $\mathcal{L}_{\mathrm{FR}}^{(l)}$ is evaluated over prefixes $\mathbf{s}_{l-1} \sim p_S$, while the training objective in Eq.~\eqref{eq:off_policy_bound} assumes $\mathbf{s}_{l-1} \sim p_T$. To formally relate them, we apply a change-of-measure expansion to $\mathcal{L}_{\mathrm{FR}}^{(l)}$:
\begin{equation}
    \begin{aligned}
        \mathcal{L}_{\mathrm{FR}}^{(l)} 
        &= \sum \nolimits_{\mathbf{s}_{l-1}} p_S(\mathbf{s}_{l-1}) D_{\mathrm{TV}} \big( \boldsymbol{\pi}_T(\cdot \mid \mathbf{s}_{l-1}) \parallel \boldsymbol{\pi}_S(\cdot \mid \mathbf{s}_{l-1}) \big) \\
        &= \mathbb{E}_{\mathbf{s}_{l-1} \sim p_T} \left[ \frac{p_S(\mathbf{s}_{l-1})}{p_T(\mathbf{s}_{l-1})} D_{\mathrm{TV}} \big( \boldsymbol{\pi}_T(\cdot \mid \mathbf{s}_{l-1}) \parallel \boldsymbol{\pi}_S(\cdot \mid \mathbf{s}_{l-1}) \big) \right].
    \end{aligned}
    \label{eq:importance_sampling}
\end{equation}
Because generation is strictly history-dependent, if the student makes an early mistake at step $k < l$, the subsequent prefix $\mathbf{s}_{l-1}$ drifts out of the teacher's data manifold. Mathematically, this implies $p_T(\mathbf{s}_{l-1}) \to 0$, causing the density ratio $\frac{p_S(\mathbf{s}_{l-1})}{p_T(\mathbf{s}_{l-1})} \to \infty$. 
Consequently, for step $l$, the expected divergence $\mathcal{L}_{\mathrm{FR}}^{(l)}$ is no longer bounded by $\epsilon$. Instead, in the worst-case out-of-distribution scenario, the divergence reaches its theoretical maximum $D_{\mathrm{TV}} \to 1$. Substituting this cascading error into Eq.~\eqref{eq:telescoping_tv} yields an expected compounding sequence error that grows quadratically $\mathcal{O}(L^2 \epsilon)$, effectively proving the \textit{exposure bias}. \hfill $\square$

\subsection{RQ2: Reversed Exposure Bias in On-Policy Distillation} \label{sec:appa.rq2}

To prevent the density ratio explosion seen in Eq.~\eqref{eq:importance_sampling}, on-policy distillation enforces the student to learn directly from its own rollout distribution: $\mathbb{E}_{\mathbf{s}_{l-1} \sim p_S} \big[ D_{\mathrm{KL}}(\boldsymbol{\pi}_S \parallel \boldsymbol{\pi}_T) \big]$. 

\textbf{Theorem 2.} \textit{Let $\mathbf{s}_{l-1} \sim p_S$ be a sub-optimal reasoning history such that it deviates fundamentally from logical coherence. Conditioning the teacher on this state causes its predictive distribution over the next step to collapse to a uniform prior, maximizing entropy. Consequently, the teacher merely provides a suppressive signal rather than constructive guidance, destroying valid supervision.}

\noindent
\textbf{Proof.}
By Bayes' theorem, the teacher's conditional probability for the next reasoning step $\mathbf{y}_l$ given the history $\mathbf{s}_{l-1}$ is defined by the joint probability of the sequence:
\begin{equation}
    \boldsymbol{\pi}_T(\mathbf{y}_l \mid \mathbf{s}_{l-1}) = \frac{p_T(\mathbf{s}_{l-1}, \mathbf{y}_l)}{p_T(\mathbf{s}_{l-1})}.
    \label{eq:bayes_teacher}
\end{equation}
In practice, LLM distributions are smoothed by a prior to prevent absolute zero probabilities. Over a reasoning step of length $K = |\mathbf{y}_l|$, the uniform prior is $U(\mathbf{y}_l) = \frac{1}{|\mathcal{V}|^K}$. We approximate the implicit generative distribution as a mixture with a smoothing factor $\lambda \in (0,1)$:
\begin{equation}
    \tilde{p}_T(\mathbf{s}_{l-1}, \mathbf{y}_l) = (1 - \lambda) p_T(\mathbf{s}_{l-1}, \mathbf{y}_l) + \lambda \frac{p_T(\mathbf{s}_{l-1})}{|\mathcal{V}|^K}. \nonumber
\end{equation}
When $\mathbf{s}_{l-1}$ is a sub-optimal prefix generated by the student, the true logical continuation density $p_T(\mathbf{s}_{l-1}, \mathbf{y}_l) \to 0$. Substituting this into the conditional distribution yields:
\begin{equation}
    \boldsymbol{\pi}_T(\mathbf{y}_l \mid \mathbf{s}_{l-1}) 
    \approx \frac{(1 - \lambda) \cdot 0 + \lambda \frac{p_T(\mathbf{s}_{l-1})}{|\mathcal{V}|^K}}{p_T(\mathbf{s}_{l-1})} \\
    = \frac{\lambda}{|\mathcal{V}|^K} \propto \frac{1}{|\mathcal{V}|^K}. \nonumber
\end{equation}
This indicates that in the sub-optimal regime, the teacher lacks valid priors, and its conditional distribution degenerates toward a uniform distribution over the sequence space. The conditional entropy of the teacher for step $l$ then diverges to:
\begin{equation}
    \mathcal{H}(\boldsymbol{\pi}_T \mid \mathbf{s}_{l-1}) = - \sum \nolimits_{\mathbf{y}_l} \boldsymbol{\pi}_T(\mathbf{y}_l \mid \mathbf{s}_{l-1}) \log \boldsymbol{\pi}_T(\mathbf{y}_l \mid \mathbf{s}_{l-1}) \longrightarrow K \log |\mathcal{V}|. \nonumber
\end{equation}
Minimizing $D_{\mathrm{KL}}(\boldsymbol{\pi}_S \parallel \boldsymbol{\pi}_T)$ against this maximum-entropy distribution implies that the teacher assigns uniformly low confidence to all possible next steps. Rather than providing active, constructive guidance (i.e., a sharp distribution peaking at the correct logical deduction), the teacher acts purely as a penalty mechanism, suppressing the student's generation across the board. This forces the student model to unlearn deterministic logical reasoning, formally defining the \textit{reversed exposure bias}. \hfill $\square$

\vspace{5pt}
\noindent
Beyond the theoretical reversed exposure bias, relying on strict on-policy supervision introduces severe practical and computational bottlenecks. 
First, calculating the token-level KL divergence fundamentally requires the teacher and student models to share the exact same tokenizer vocabulary $\mathcal{V}$ \citep{yan2026distribution}. This constraint severely restricts the flexibility of cross-architecture distillation.
Second, on-policy distillation forces the massive teacher model to synchronously evaluate or generate target distributions conditioned on the student's real-time rollouts. This step-by-step dependency introduces prohibitive computational latency and memory overhead, rendering large-scale on-policy distillation highly inefficient in practice.

\subsection{RQ3: The Failure of Naive Distribution Mixing}  \label{sec:appa.rq3}

Given the complementary flaws of off-policy and on-policy distillation, an intuitive baseline would be to optimize the student over a mixture distribution, \eg SKD \citep{xu2025speculative}, or combine the forward and reversed KL divergences \citep{jung2025todi}. Let $p_{\text{mix}} = \alpha p_S + (1-\alpha) p_T$ be a linearly mixed historical context distribution with a mixture coefficient $\alpha \in (0,1)$. The expected distillation objective becomes:
\begin{equation}
    \mathcal{J}_{\text{mix}}(\boldsymbol{\pi}_S) = \mathbb{E}_{\mathbf{s}_{l-1} \sim p_{\text{mix}}} \left[ D_{\mathrm{KL}} \big( \boldsymbol{\pi}_T(\cdot \mid \mathbf{s}_{l-1}) \parallel \boldsymbol{\pi}_S(\cdot \mid \mathbf{s}_{l-1}) \big) \right]. \nonumber
\end{equation}

\textbf{Theorem 3 (zero-sum trade-off).} \textit{Optimizing over a naive mixture distribution $p_{\text{mix}}$ cannot concurrently resolve the dual exposure biases. The expected inference error is strictly lower-bounded by a convex combination of the forward error cascade (Theorem 1) and the reversed entropy collapse (Theorem 2). Consequently, any choice of $\alpha$ merely trades one form of bias for another.}

\noindent
\textbf{Proof.}
By expanding the definition of the expectation over the discrete state space, we substitute the definition of the mixed density $p_{\text{mix}}(\mathbf{s}_{l-1}) = \alpha p_S(\mathbf{s}_{l-1}) + (1-\alpha) p_T(\mathbf{s}_{l-1})$ into the objective:
\begin{equation}
    \begin{aligned}
        \mathcal{J}_{\text{mix}} &= \sum \nolimits_{\mathbf{s}_{l-1}} p_{\text{mix}}(\mathbf{s}_{l-1}) D_{\mathrm{KL}} \big( \boldsymbol{\pi}_T(\cdot \mid \mathbf{s}_{l-1}) \parallel \boldsymbol{\pi}_S(\cdot \mid \mathbf{s}_{l-1}) \big) \\
        &= \sum \nolimits_{\mathbf{s}_{l-1}} \big[ \alpha p_S(\mathbf{s}_{l-1}) + (1-\alpha) p_T(\mathbf{s}_{l-1}) \big] D_{\mathrm{KL}} \big( \boldsymbol{\pi}_T(\cdot \mid \mathbf{s}_{l-1}) \parallel \boldsymbol{\pi}_S(\cdot \mid \mathbf{s}_{l-1}) \big). \nonumber
    \end{aligned}
\end{equation}
By distributing the summation, the mixture objective $\mathcal{J}_{\text{mix}}$ is explicitly decoupled into two independent expectations:
\begin{equation}
    \mathcal{J}_{\text{mix}} = \alpha \underbrace{ \mathbb{E}_{\mathbf{s}_{l-1} \sim p_S} \left[ D_{\mathrm{KL}} \big( \boldsymbol{\pi}_T \parallel \boldsymbol{\pi}_S \big) \right] }_{\text{On-policy constraint}} + (1-\alpha) \underbrace{ \mathbb{E}_{\mathbf{s}_{l-1} \sim p_T} \left[ D_{\mathrm{KL}} \big( \boldsymbol{\pi}_T \parallel \boldsymbol{\pi}_S \big) \right] }_{\text{Off-policy constraint}}. \nonumber
\end{equation}
This exact decoupling reveals a structural contradiction between the two terms.

First, consider the $\alpha$ component. For historical prefixes sampled directly from the student's rollout ($\mathbf{s}_{l-1} \sim p_S$), the trajectories inevitably contain out-of-distribution trajectories where $p_T(\mathbf{s}_{l-1}) \to 0$. As rigorously established in Theorem 2, the teacher's conditional entropy diverges. Thus, minimizing the KL divergence forces the student to fit a maximum-entropy distribution, inflicting a supervision penalty proportional to $\mathcal{H}(\boldsymbol{\pi}_T \mid \mathbf{s}_{l-1}) \to K \log |\mathcal{V}|$.

Second, consider the $(1-\alpha)$ component. To avoid the aforementioned entropy collapse, one must heavily weight the off-policy data by setting $\alpha \to 0$. However, as proven in Theorem 1, optimizing exclusively over $p_T$ leaves the student vulnerable to compounding sequence-level errors during test-time free-running generation. This inference error cascade is mathematically bounded by the trajectory-level divergence $D_{\mathrm{TV}} \big( p_T(\mathbf{s}_L) \parallel p_S(\mathbf{s}_L) \big) \ge \Omega(L^2 \epsilon)$.

Formally, we define the total sub-optimality risk functional $\mathcal{R}_{\text{total}}$ as the weighted sum of the expected training supervision corruption and the expected inference sequence deviation. The risk is strictly bounded from below:
\begin{equation}
    \mathcal{R}_{\text{total}} \ge \alpha \cdot \mathbb{E}_{\mathbf{s}_{l-1} \sim p_S} \big[ \mathcal{H}(\boldsymbol{\pi}_T \mid \mathbf{s}_{l-1}) \big] + (1-\alpha) \cdot D_{\mathrm{TV}} \big( p_T(\mathbf{s}_L) \parallel p_S(\mathbf{s}_L) \big). \nonumber
\end{equation}
Because the two penalty terms are mutually antagonistic, lowering $\alpha$ to secure valid supervision inherently amplifies the sequential divergence, and raising $\alpha$ to cover the student's distribution actively corrupts the reasoning targets. Thus, the structural deadlock remains unbroken. The theoretical oracle requires valid supervision \textit{conditioned on} sub-optimal contexts, a property that naive distribution mixing mathematically fails to provide. \hfill $\square$

\subsection{RQ4: Theoretical Guarantees of \baby}  \label{sec:appa.rq4}

We formally prove how \baby satisfies both the \textit{coverage} and \textit{validity} conditions. Let $p_{\boldsymbol{\pi}_S}^\gamma$ denote the truncated distribution of prefixes generated under the \baby adaptive boundary mechanism.

\textbf{Theorem 4 (coverage).} \textit{The truncated distribution $p_{\boldsymbol{\pi}_S}^\gamma$ strictly covers the authentic student distribution $p_S$ up to a tightly bounded divergence, eliminating the exposure bias without exposing the teacher to fatal logical collapses.}

\noindent
\textbf{Proof.}
The synthesized distribution $p_{\boldsymbol{\pi}_S}^\gamma$ matches $p_S$ until a truncation occurs, triggered when $V_T(\mathbf{s}_{l-1}, \mathbf{y}_l^S) < \gamma(\mathbf{s}_{l-1})$. The distance between the true student distribution and the truncated distribution is bounded by the expected cumulative probability of the truncation over the $L$ steps:
\begin{equation}
    D_{\mathrm{TV}}(p_S \parallel p_{\boldsymbol{\pi}_S}^\gamma) \le \sum \nolimits_{l=1}^L \mathbb{P}_{\mathbf{s}_{l-1} \sim p_S, \mathbf{y}_l \sim \boldsymbol{\pi}_S} \left( V_T(\mathbf{s}_{l-1}, \mathbf{y}_l) < \gamma(\mathbf{s}_{l-1}) \right).
    \label{eq:coverage_bound}
\end{equation}
Recall our adaptive boundary formulation: 
\begin{equation}
    \gamma(\mathbf{s}_{l-1}) = \gamma_0 \exp \big(-\alpha \mathcal{H}(\boldsymbol{\pi}_T \mid \mathbf{s}_{l-1}) \big). \nonumber
\end{equation}
When the teacher identifies multiple valid reasoning paths, \ie high entropy $\mathcal{H}$, $\gamma(\mathbf{s}_{l-1})$ exponentially decays. This ensures the condition $V_T < \gamma$ is \textit{strictly conservative} and triggers only when $V_T \to 0$ (an irrecoverable logical error), rather than for stylistic variances. 
Since $\gamma(\mathbf{s}_{l-1}) \le \gamma_0$, the total variation in Eq.~\eqref{eq:coverage_bound} is rigorously constrained by the base scalar $\gamma_0$. Therefore, $p_{\boldsymbol{\pi}_S}^\gamma \approx p_S$ across recoverable reasoning manifolds, verifying the \textit{coverage condition}. \hfill $\square$

\textbf{Theorem 5 (validity).} \textit{\baby unconditionally prevents the entropy collapse described in Theorem 2, providing valid oracle supervision and fully circumventing the reversed exposure bias.}

\noindent
\textbf{Proof.}
In \baby, upon detecting a boundary breach at step $l$, the framework strictly avoids querying $\boldsymbol{\pi}_T(\cdot \mid \mathbf{s}_l)$. Instead, credit assignment backtracks to a previous step $l^*$, and the off-policy target $\mathcal{Y}^T$ is sampled recursively as:
\begin{equation}
    \mathcal{Y}^T_{l^*:L} \sim \prod \nolimits_{\tau=l^*}^L \boldsymbol{\pi}_T(\mathbf{y}_\tau \mid \mathbf{s}_{\tau-1}^T), \quad \text{where } \mathbf{s}_{l^*-1}^T = \mathbf{s}_{l^*-1}. \nonumber
\end{equation}
Because the state $\mathbf{s}_{l^*-1}$ was verified as safe prior to backtracking, it satisfies the boundary constraint $V_T(\mathbf{s}_{l^*-2}, \mathbf{y}_{l^*-1}) \ge \gamma(\mathbf{s}_{l^*-2})$. 
This guarantees a high transition likelihood, ensuring the root context remains strictly within the teacher's dense data manifold:
\begin{equation}
    p_T(\mathbf{s}_{l^*-1}) \ge \delta \gg 0. \nonumber
\end{equation}
Following the derivation in Eq.~\eqref{eq:bayes_teacher}, because $p_T(\mathbf{s}_{l^*-1})$ is substantially strictly positive, the conditional distribution $\boldsymbol{\pi}_T(\cdot \mid \mathbf{s}_{l^*-1})$ remains sharp and highly confident. The step-level entropy $\mathcal{H}(\boldsymbol{\pi}_T \mid \mathbf{s}_{l^*-1})$ is thus bounded securely away from the uniform maximum $K \log |\mathcal{V}|$.

Finally, the asynchronous stitching objective constructs the loss over the \texttt{[REV]} semantic transition:
\begin{equation}
    \mathcal{L}_{\text{stitch}} = \mathbb{E}_{\mathcal{Y}^T} \left[ -\log \boldsymbol{\pi}_S \big(\texttt{[REV]} \oplus \mathcal{Y}^T \mid \mathbf{s}_l^\text{flawed} \big) \right]. \nonumber
\end{equation}
This isolates the teacher's generation process from the flawed context $\mathbf{s}_l^\text{flawed}$. The teacher operates purely on the pristine root $\mathbf{s}_{l^*-1}$, while the student learns a conditional mapping from its own flawed exploration to a pristine corrective sequence via the \texttt{[REV]} bottleneck. Consequently, the teacher's supervision signal remains unconditionally valid, neutralizing the reversed exposure bias. \hfill $\square$

\section{Algorithmic Pipeline of \baby} \label{app:algorithm}

In this section, we present the detailed algorithmic procedure of our proposed data synthesis framework, \baby. The process is iteratively executed over a set of unlabelled reasoning questions to construct a refined supervised fine-tuning dataset. 

As outlined in Alg.~\ref{alg:baby}, the pipeline requires a teacher policy $\boldsymbol{\pi}_T$, a student policy $\boldsymbol{\pi}_S$, and an initial set of questions $\mathcal{Q}$. For each question, the student auto-regressively generates reasoning steps while the teacher synchronously evaluates the transition likelihood $V_T$ and context entropy $\mathcal{H}$. If the generation remains within the adaptive safety boundary $\gamma$, the step is accepted, preserving the student's authentic on-policy distribution. Once a critical logical drift is detected ($V_T < \gamma$), the algorithm triggers the credit assignment mechanism to trace back the temporal difference errors $\delta_k$. It identifies the optimal safe bifurcation state $\mathbf{s}_{l^*-1}$, prompts the teacher to generate a pristine correction $\mathcal{Y}^T_{l^*:T}$, and stitches the flawed exploration with the oracle correction using a specific \texttt{[REV]} token. The resulting trajectories explicitly empower the student model with both robustness to its own stylistic variances and the capability for logical self-correction.

\begin{algorithm}[t]
\caption{Complete pipeline of the data synthesis via \baby}
\label{alg:baby}
\textbf{Input:} Question set $\mathcal{Q}$, Teacher policy $\boldsymbol{\pi}_T$, Student policy $\boldsymbol{\pi}_S$, Base threshold $\gamma_0$, Scaling factor $\alpha$. \\
\textbf{Output:} Synthesized supervised fine-tuning dataset $\mathcal{D}_\text{SFT}$.
\begin{algorithmic}[1]
\STATE Initialize dataset $\mathcal{D}_\text{SFT} \leftarrow \varnothing$;
\FOR{each question $\mathbf{q} \in \mathcal{Q}$}
    \STATE Initialize context $\mathbf{s}_0 \leftarrow \mathbf{q}$, step index $l \leftarrow 1$;
    \STATE Flag $is\_safe \leftarrow \text{True}$;
    \WHILE{$is\_safe$ \AND not reached terminal state}
        \STATE Sample student reasoning step: $\mathbf{y}_l^S \sim \boldsymbol{\pi}_S(\cdot \mid \mathbf{s}_{l-1})$;
        \STATE Compute step value $V_T(\mathbf{s}_{l-1}, \mathbf{y}_l^S)$ via Eq.~\eqref{eq:value_function};
        \STATE Compute teacher entropy $\mathcal{H}(\boldsymbol{\pi}_T \mid \mathbf{s}_{l-1})$;
        \STATE Calculate adaptive boundary $\gamma(\mathbf{s}_{l-1}) \leftarrow \gamma_0 \cdot \exp(-\alpha \cdot \mathcal{H})$;
        
        \IF{$V_T(\mathbf{s}_{l-1}, \mathbf{y}_l^S) \ge \gamma(\mathbf{s}_{l-1})$}
            \STATE \textit{\color{Blue} $\triangleright$ Safe step, continue on-policy unrolling}
            \STATE Update context $\mathbf{s}_l \leftarrow [\mathbf{s}_{l-1}, \mathbf{y}_t^S]$;
            \STATE $l \leftarrow l + 1$;
        \ELSE
            \STATE \textit{\color{Blue} $\triangleright$ Boundary breached, trigger backtracking}
            \STATE $is\_safe \leftarrow \text{False}$;
            \FOR{$k = 1$ \TO $t$}
                \STATE Compute temporal difference error $\delta_k = V_T(\mathbf{s}_{k-1}, \mathbf{y}_k^S) - V_T(\mathbf{s}_{k-2}, \mathbf{y}_{k-1}^S)$;
            \ENDFOR
            \STATE Identify optimal safe root $l^* \leftarrow \arg\min_{k \le l} (\delta_k)$ s.t. prior state is safe;
            \STATE \textit{\color{Blue} $\triangleright$ Off-policy correction and stitching}
            \STATE Sample corrective trajectory $\mathcal{Y}_{l^*:T}^T \sim \boldsymbol{\pi}_T(\cdot \mid \mathbf{s}_{l^*-1})$;
            \STATE Construct stitched trajectory $\boldsymbol{\tau} \leftarrow [\mathbf{q}, \mathcal{Y}_{1:l}^S] \oplus \texttt{[REV]} \oplus \mathcal{Y}_{l^*:T}^T$;
            \STATE $\mathcal{D}_\text{SFT} \leftarrow \mathcal{D}_\text{SFT} \cup \{\boldsymbol{\tau}\}$;
        \ENDIF
    \ENDWHILE
    \IF{$is\_safe$}
        \STATE \textit{\color{Blue} $\triangleright$ Successfully finished without correction}
        \STATE $\mathcal{D}_\text{SFT} \leftarrow \mathcal{D}_\text{SFT} \cup \{\mathbf{s}_l\}$;
    \ENDIF
\ENDFOR
\STATE \textbf{return} $\mathcal{D}_\text{SFT}$
\end{algorithmic}
\end{algorithm}

\section{Computational Complexity Analysis}
\label{subsec:complexity}

In addition to resolving the theoretical dual exposure biases, \baby introduces significant computational advantages during the data synthesis pipeline. In this section, we formally analyze the time complexity of \baby and compare it against the SKD baseline introduced in Sec.~\ref{sec:gap}.

To formulate the computational overhead, let $\Lambda$ denote the total token length of a reasoning trajectory. For an LLM parameterized by $\boldsymbol{\theta}$, we distinguish between two primary operations: 
(1) \textit{Auto-regressive generation} ($\mathcal{O}_\text{gen}(\boldsymbol{\theta}, \Lambda)$), which requires sequential decoding and saving KV cache, which is memory-bandwidth bound; 
(2) \textit{Parallel verification and forward pass} ($\mathcal{O}_\text{vfy}(\boldsymbol{\theta}, \Lambda)$), which computes logits over a given sequence in parallel. 
Due to parallelized KV-caching and matrix multiplications, verification is significantly faster than generation, \ie, $\mathcal{O}_\text{vfy} \ll \mathcal{O}_\text{gen}$.

\noindent \textbf{Complexity of SKD.}
In standard SKD pipelines, the teacher model $\boldsymbol{\pi}_T$ continuously monitors the student $\boldsymbol{\pi}_S$ at the token or granular step level. Whenever the student's logic slightly deviates, SKD forces the teacher to intervene and generate a replacement. Assuming an intervention rate $\alpha \in (0, 1)$, the total time complexity per trajectory is:
\begin{equation}
    \mathcal{O}_\text{SKD} \approx \mathcal{O}_\text{gen}(\boldsymbol{\theta}_S, \Lambda) + \mathcal{O}_\text{vfy}(\boldsymbol{\theta}_T, \Lambda) + \alpha \cdot \mathcal{O}_\text{gen}(\boldsymbol{\theta}_T, \Lambda). \nonumber
\end{equation}
Because SKD enforces strict alignment, the intervention rate $\alpha$ is inherently high. The frequent context switching and repeated auto-regressive generation by the massively parameterized teacher ($\boldsymbol{\theta}_T > \boldsymbol{\theta}_S$) result in a severe computational bottleneck.

\noindent \textbf{Complexity of \baby.}
Conversely, \baby isolates operations to minimize the teacher's generation burden. Assuming a boundary breach occurs at step $l$ (corresponding to token length $\Lambda_\text{fail}$), and the safe point is identified at step $l^*$, the complexity of \baby is decomposed as follows:
\begin{itemize}
    \item \textbf{Phase 1 (monitoring):} The student freely generates the prefix, and the teacher evaluates the steps using Eq.~\eqref{eq:value_function} and \eqref{eq:adaptive_gamma}. Because the student's text is already generated, the teacher performs highly efficient parallel forward passes. The cost is $\mathcal{O}_\text{gen}(\boldsymbol{\theta}_S, \Lambda_\text{fail}) + \mathcal{O}_\text{vfy}(\boldsymbol{\theta}_T, \Lambda_\text{fail})$.
    \item \textbf{Phase 2 (backtracking):} Identifying the root cause $l^*$ relies on computing the temporal difference error $\delta_k$. Since the step-wise values $V_T$ are already cached during Phase 1, this operation requires zero neural network passes. It is purely algebraic, incurring a negligible complexity of $\mathcal{O}(1)$.
    \item \textbf{Phase 3 (stitching):} The teacher generates the corrective suffix of length $\Lambda_\text{correct}$ from the root cause. The cost is $\mathcal{O}_\text{gen}(\boldsymbol{\theta}_T, \Lambda_\text{correct})$.
\end{itemize}
Summing these phases, the total time complexity for \baby is:
\begin{equation}
    \mathcal{O}_{\text{\baby}} \approx \mathcal{O}_\text{gen}(\boldsymbol{\theta}_S, \Lambda_\text{fail}) + \mathcal{O}_\text{vfy}(\boldsymbol{\theta}_T, \Lambda_\text{fail}) + \mathcal{O}_\text{gen}(\boldsymbol{\theta}_T, \Lambda_\text{correct}).
\end{equation}
The fundamental efficiency advantage of \baby lies in the adaptive boundary. By tolerating sub-optimal explorations, \baby drastically reduces the necessity for teacher interventions, shifting the teacher's computational workload from expensive auto-regressive generation ($\mathcal{O}_\text{gen}$) to highly efficient parallel verification ($\mathcal{O}_\text{vfy}$). Furthermore, unlike external verification methods, \eg process reward models, that require training an additional critic network, our backtracking by temporal difference error reuses the teacher's innate confidence, effectively achieving precise credit assignment at zero additional inference cost.

\noindent \textbf{Empirical computational budgets.}
To validate the practical efficiency of \baby, we measure the real-world data synthesis speed on a hardware setup consisting of 8$\times$A800 GPUs. We compare the average generation time of \baby against the SKD baseline across three distinct hyperparameter configurations (which control the intervention rates by the teacher). Empirically, both methods demonstrate comparable synthesis efficiency, each requiring approximately 3 to 4 hours to generate 1k reasoning trajectories on average. This result confirms that while \baby introduces sophisticated mechanisms, \eg backtracking with temporal difference error and trajectory stitching, to theoretically resolve the dual exposure biases, it does not incur noticeable empirical overhead in practice.

\section{Detailed Experimental Settings}\label{app.c}

In this section, we provide a comprehensive overview of the experimental setups, including the specific datasets, model architectures, baseline methodologies, and implementation details.

\subsection{Training Datasets}

Our training phase leverages high-quality reasoning problem sets to facilitate effective knowledge distillation. Specifically, the training corpus is constructed from two primary sources:
\begin{itemize}
    \item \textbf{\textit{LIMO-v2}} \citep{ye2025limo}: We utilize problem sets from the \textit{LIMO-v2} repository\footnote{\url{https://huggingface.co/datasets/GAIR/LIMO-v2}.}, which is specifically curated to elicit deep mathematical reasoning capabilities. This dataset has undergone rigorous quality filtering and comprises 800 mathematical problems, and we generate five distinct responses for each problem.
    \item \textbf{\textit{AceReason}} \citep{liu2025acereason}: To introduce further diversity and scale to our training distribution, we randomly sample a subset of 10,000 problem instances from the \textit{AceReason} repository\footnote{\url{https://huggingface.co/datasets/nvidia/AceReason-1.1-SFT}.}. For each problem, we generate one response.
\end{itemize}

\subsection{Evaluation Benchmarks}

To rigorously assess the generalization and reasoning capability of our trained student models, we employ a suite of five challenging benchmarks, categorized into domain-specific mathematical reasoning and cross-domain capabilities:
\begin{itemize}
    \item \textbf{Mathematical reasoning:} We evaluate our student models on \textit{AIME2024} and \textit{AIME2025}, which consist of highly complex, competition-level problems from the American Invitational Mathematics Examination. Each test set comprises 30 challenging mathematical problems. During evaluation, we generate 8 responses per problem and report the pass@$k$ accuracy, where at least one correct response is identified among $k$ samples, for $k \in \{8, 4, 1\}$. 
    Additionally, we use \textit{MATH500} \citep{lightman2024let}, a comprehensive benchmark encompassing various difficulty levels and subfields of mathematics. We evaluate the student models across varying difficulty levels, ranging from Level 1 to Level 5. The respective test sets consist of 43, 90, 105, 128, and 134 problems, for which we report accuracy scores individually.
    \item \textbf{Cross-domain capabilities:} Our models are trained on two mathematical reasoning datasets. To ensure that our approach generalizes to other domains, \eg general instruction-following and diverse knowledge reasoning, we evaluate our models on the \textit{IFEval} benchmark \citep{zhou2023instruction}, which assesses rigorous, verifiable instruction adherence. This benchmark comprises 541 prompts. We report both the widely used prompt-level-strict accuracy and the mean across all four metrics, including prompt-level-strict, inst-level-strict, prompt-level-loose, and inst-level-loose. Furthermore, we evaluate our models on \textit{GPQA} \citep{rein2023gpqa}, a challenging graduate-level benchmark covering physics, biology, and chemistry. This benchmark consists of 198 challenging scientific questions. Consistent with our evaluation protocol on the AIME datasets, we perform eight samplings per problem and report the pass@k metrics.
\end{itemize}

\subsection{Model Architectures}

We evaluate our method across LLMs of varying scales and architectural families.

\noindent
\textbf{Teacher LLMs.} 
We employ \textit{Qwen3-32B}\footnote{\url{https://huggingface.co/Qwen/Qwen3-32B}.} and \textit{gpt-oss-120b}\footnote{\url{https://huggingface.co/openai/gpt-oss-120b}.} as our teacher models. These large-scale models possess extensive parameter counts and demonstrate state-of-the-art proficiency in generating high-quality, accurate, and complex reasoning trajectories.

\noindent
\textbf{Student LLMs.} 
To evaluate the efficacy and scalability of our method, we select \textit{Qwen3-4B-Instruct}\footnote{\url{https://huggingface.co/Qwen/Qwen3-4B-Instruct-2507}.}, \textit{Qwen2.5-7B-Instruct}\footnote{\url{https://huggingface.co/Qwen/Qwen2.5-7B-Instruct}.}, and \textit{Llama3.2-3B-Instruct}\footnote{\url{https://huggingface.co/meta-llama/Llama-3.2-3B-Instruct}.} as our student models. These models represent lightweight, efficient architectures (ranging from 3B to 7B parameters) that are highly suitable for real-world deployments requiring low-latency inference.

\subsection{Baseline Methods}

We compare our proposed approach against four representative baselines spanning standard fine-tuning and knowledge distillation paradigms.

\begin{itemize}
    \item \textbf{Supervised fine-tuning (SFT):} The student model is directly fine-tuned on reasoning trajectories generated by the teacher, serving as a fundamental reference baseline.
    \item \textbf{Rejection sampling fine-tuning (RFT):} Multiple candidate trajectories are sampled from the student model; those leading to incorrect final answers are discarded, and the student is fine-tuned exclusively on the surviving correct rollouts.
    \item \textbf{Imitation-based knowledge distillation (ImitKD)}~\citep{lin2020autoregressive}: An autoregressive imitation learning method in which, at each decoding step, the generation is stochastically sourced from either the teacher or the student.
    \item \textbf{Speculative knowledge distillation (SKD)}~\citep{xu2025speculative}: A step-level, speculative distillation scheme. At each reasoning step $l$, the student first generates a candidate step $\mathbf{a}^S_l$, which is then evaluated by both the teacher and the student via their respective geometric mean token probabilities:
    \begin{equation}
        \bar{p}_M\!\left(\mathbf{a}^S_l \mid \mathbf{s}_{l-1}\right)
        = \exp\!\left(\frac{1}{\left|\mathbf{a}^S_l\right|}\sum \nolimits_{i=1}^{\left|\mathbf{a}^S_l\right|}
          \log p_M\!\left(y_i \mid \mathbf{s}_{l-1}, \mathbf{y}_{<i}\right)\right), \quad M \in \{S, T\}. \nonumber
    \end{equation}
    The student's step is accepted if the teacher's likelihood sufficiently dominates the student's:
    \begin{equation}
        \bar{p}_T\!\left(\mathbf{a}^S_l \mid \mathbf{s}_{l-1}\right)
        \;>\;
        \beta \cdot \bar{p}_S\!\left(\mathbf{a}^S_l \mid \mathbf{s}_{l-1}\right), \nonumber
    \end{equation}
    where $\beta \in (0, 1]$ is a margin factor controlling acceptance strictness. When this condition fails, the student's step is discarded, and the teacher regenerates a replacement, steering the trajectory back toward high-quality reasoning.
\end{itemize}

Meanwhile, we also compare several recent on-policy distillation methods, which are described as follows:

\begin{itemize}
    \item \textbf{Top-$k$ on-policy distillation (top-$k$ OPD)}~\citep{fu2026revisiting}: This method mitigates the train-inference distribution mismatch by training the student model on its own self-generated sequences rather than a fixed dataset. It leverages token-level teacher probabilities as expert feedback on these student outputs, accommodating flexible divergence measures and seamless integration with reinforcement learning. Unlike GKD \citep{agarwal2024on}, which calculates the divergence over the entire vocabulary-level distribution, this method utilizes the top-$k$ tokens to enhance computational efficiency.
    \item \textbf{Distribution-aligned sequence distillation (DASD)}~\citep{yan2026distribution}: Designed to enhance sequence-level distillation for long chain-of-thought reasoning, this method integrates temperature-scheduled learning and divergence-aware sampling to better cover and align with the teacher's distribution. To explicitly combat exposure bias, it employs a mixed-policy strategy that corrects student-generated prefixes with targeted teacher continuations.
\end{itemize}

\subsection{Implementation Details}

\noindent
\textbf{Data generation.}
All models are served via SGLang. During the data generation pipeline of \baby, the student generates reasoning steps delimited by the stop sequence \texttt{".\textbackslash n\textbackslash n"}, with each step capped at 8,192 tokens and the full trajectory bounded at 32,768 tokens.
The teacher's predictive entropy $\mathcal{H}(\boldsymbol{\pi}_T \mid \mathbf{s}_{t-1})$ is estimated from the renormalized top-20 next-token logprobs.
Upon an unsafe point, the backtracking based on the temporal difference error identifies the point $l^*$ (subject to the constraint that step $k{-}1$ was still above its adaptive threshold), after which the teacher generates a full corrective suffix from the pristine context $s_{l^*-1}$, and the final SFT trajectory is assembled as: student exploration $+$ \texttt{``However,''} $+$ teacher correction.
We set $\gamma_0 \in \{0.1, 0.2, 0.3, 0.4, 0.5\}$ and temperature $\tau = 0.6$ for Qwen3-32B or $1.0$ for gpt-oss-120b, and generate $N=5$ trajectories per question in \textit{LIMO-v2} dataset and $N=1$ for \textit{AceReason}.

\noindent
\textbf{Student model training.}
The student models are trained with full-parameter SFT using LLaMA-Factory on 8 GPUs via \texttt{torchrun}, with \textit{DeepSpeed ZeRO-Stage~3}, \textit{Flash Attention~2}, and the \textit{Liger Kernel} enabled.
To accelerate training, we employ sequence packing, where shorter training samples are concatenated into sequences of up to 32,768 tokens to maximize context utilization.
We train for 6 epochs with a per-device batch size of~4 (effective global batch size 32), using AdamW with $\beta_1{=}0.9$, $\beta_2{=}0.95$, $\epsilon{=}10^{-8}$, weight decay $0.1$, and gradient clipping at norm $1.0$.
The learning rate is $5{\times}10^{-5}$ with a cosine decay schedule (minimum $10^{-6}$) and a linear warmup over 5\% of total steps.

\begin{table}[t]
\centering
\renewcommand{\arraystretch}{1.10}
\setlength{\tabcolsep}{4pt}
\caption{Complete experimental results on more student models using Qwen3-32B as the teacher model. The bold results indicate the best scores.}
\label{table:app.more_student}
\small
    \begin{tabular}{m{1.85cm}<{\centering}m{0.65cm}<{\centering}m{0.65cm}<{\centering}m{0.65cm}<{\centering}m{0.65cm}<{\centering}m{0.65cm}<{\centering}m{0.65cm}<{\centering}m{0.65cm}<{\centering}m{0.65cm}<{\centering}m{0.65cm}<{\centering}m{0.65cm}<{\centering}m{0.65cm}<{\centering}m{0.69cm}<{\centering}m{0.69cm}<{\centering}}
    \toprule
    \multirow{2}{*}{Method} & \multicolumn{3}{c}{AIME24} & \multicolumn{3}{c}{AIME25} & \multicolumn{5}{c}{MATH500} & IFEval & GPQA \\
    \cmidrule(r){2-4} \cmidrule(r){5-7} \cmidrule(r){8-12} \cmidrule(r){13-13} \cmidrule(r){14-14}
    & P@8 & P@4 & P@1 & P@8 & P@4 & P@1 & L1 & L2 & L3 & L4 & L5 & Avg. & P@8 \\
    \midrule
    \multicolumn{14}{c}{\textbf{Student LLM}: Qwen3-1.7B} \\
    SFT & 36.67 & 25.90 & 9.58 & 26.67 & 22.29 & 11.25 & \textbf{90.70} & 84.44 & 73.33 & 67.97 & 43.28 & 43.38 & 68.18 \\
    RFT & 13.33 & 11.38 & 5.83 & 33.33 & 24.67 & 12.08 & 81.40 & 75.56 & 74.29 & 66.41 & 41.04 & 41.02 & 60.10 \\
    ImitKD & 36.67 & 30.19 & 16.25 & 36.67 & 28.33 & 12.50 & 81.40 & 80.00 & 80.00 & 59.38 & 48.51 & 43.31 & 63.13 \\
    SKD {\footnotesize ($\beta=0.6$)} & 23.33 & 17.57 & 7.92 & 26.67 & 22.33 & 9.59 & \textbf{90.70} & 83.33 & 77.14 & 66.41 & 45.52 & 44.36 & 61.62 \\
    SKD {\footnotesize ($\beta=0.8$)} & 43.33 & 34.71 & 14.58 & 36.67 & 28.24 & 13.33 & 88.37 & 85.56 & 76.19 & 68.75 & 44.03 & 44.36 & \textbf{73.74} \\
    SKD {\footnotesize ($\beta=1.0$)} & 23.33 & 18.57 & 10.83 & 33.33 & 23.90 & 9.17 & 86.05 & 76.67 & 69.52 & 60.16 & 41.04 & 44.19 & 62.63 \\
    \hline
    \rowcolor{lightgrayv} \textbf{\baby} {\footnotesize (ours)} & \textbf{50.00} & \textbf{41.38} & \textbf{22.08} & \textbf{40.00} & \textbf{32.05} & \textbf{14.58} & \textbf{90.70} & \textbf{93.33} & \textbf{82.86} & \textbf{78.91} & \textbf{58.21} & \textbf{46.80} & \textbf{73.74} \\
    \hline
    \specialrule{0em}{0.5pt}{0.5pt}
    \hline
    \multicolumn{14}{c}{\textbf{Student LLM}: Llama3.2-3B-Instruct} \\
    SFT & 16.67 & 11.19 & 3.33 & 6.67 & 3.33 & 0.83 & 79.07 & 53.33 & 51.43 & 42.19 & 22.39 & 42.30 & 58.59 \\
    RFT & 10.00 & 7.38 & 2.50 & 6.67 & 3.33 & 0.83 & 74.42 & 58.89 & 59.05 & 42.19 & 20.90 & 41.16 & 53.54 \\
    ImitKD & 16.67 & 9.76 & 2.92 & 6.67 & 4.76 & 1.67 & 76.74 & 66.67 & 57.14 & 40.62 & 17.91 & 43.35 & 66.67 \\
    SKD {\footnotesize ($\beta=0.6$)} & 16.67 & 8.33 & 2.08 & 6.67 & 4.95 & 2.08 & 76.64 & \textbf{68.89} & 58.10 & 42.97 & 15.67 & 41.57 & 64.65 \\
    SKD {\footnotesize ($\beta=0.8$)} & 20.00 & 11.67 & 2.92 & 13.33 & 9.05 & 2.92 & \textbf{88.37} & 63.33 & 59.05 & \textbf{50.00} & 23.88 & 41.65 & 67.68 \\
    SKD {\footnotesize ($\beta=1.0$)} & 16.67 & 12.38 & 3.75 & 6.67 & 5.24 & 1.67 & 83.72 & 58.89 & 52.38 & 34.38 & 17.91 & 39.79 & 71.21 \\
    \hline
    \rowcolor{lightgrayv} \textbf{\baby} {\footnotesize (ours)} & \textbf{23.33} & \textbf{13.76} & \textbf{5.00} & \textbf{16.67} & \textbf{11.67} & \textbf{5.83} & 86.05 & \textbf{68.89} & \textbf{60.95} & 48.44 & \textbf{28.36} & \textbf{47.67} & \textbf{73.74} \\
    \hline
    \specialrule{0em}{0.5pt}{0.5pt}
    \hline
    \multicolumn{14}{c}{\textbf{Student LLM}: Qwen2.5-7B-Instruct} \\
    SFT & 23.33 & 19.76 & \textbf{10.42} & 16.67 & 10.00 & 4.58 & 83.72 & 74.44 & 69.52 & 60.16 & 44.78 & 34.78 & 69.70 \\
    RFT & 16.67 & 12.86 & 5.42 & 20.00 & 14.00 & 5.00 & 88.37 & 82.22 & 80.95 & 58.59 & 35.07 & 34.27 & 56.06 \\
    ImitKD & 20.00 & 16.19 & 5.83 & \textbf{26.67} & 16.57 & 5.83 & 74.42 & 70.00 & 72.38 & 57.81 & 35.82 & 35.32 & 70.71 \\
    SKD {\footnotesize ($\beta=0.6$)} & 23.33 & 19.24 & 10.00 & 23.33 & 17.86 & \textbf{7.92} & 83.72 & 83.33 & 75.24 & 65.62 & 33.58 & 36.18 & 70.71 \\
    SKD {\footnotesize ($\beta=0.8$)} & 26.67 & 22.05 & 10.00 & \textbf{26.67} & 18.81 & 7.50 & 88.37 & 81.11 & 73.33 & \textbf{69.53 }& 35.07 & 33.85 & 76.77 \\
    SKD {\footnotesize ($\beta=1.0$)} & 20.00 & 13.33 & 4.17 & 23.33 & 17.14 & 6.67 & 86.05 & 77.78 & 72.38 & 64.06 & 41.79 & 34.63 & 77.27 \\
    \hline
    \rowcolor{lightgrayv} \textbf{\baby} {\footnotesize (ours)} & \textbf{30.00} & \textbf{23.33} & \textbf{10.42} & \textbf{26.67} & \textbf{19.52} & \textbf{7.92} & \textbf{90.70} & \textbf{84.44} & \textbf{81.90} & 67.19 & \textbf{48.51} & \textbf{37.61} & \textbf{79.29} \\
    \bottomrule
    \end{tabular}
\end{table}

\noindent
\textbf{Implementation of on-policy distillation.}
We distill from a Qwen3-8B teacher into a Qwen3-1.7B student using the framework from \citet{fu2026revisiting}. The three components of our on-policy pipeline: \textit{rollout generation}, \textit{teacher forward pass}, and \textit{student optimization}, execute in a time-sliced fashion on the same set of GPUs, with each component activated only during its designated phase to accommodate all three models within the available memory budget.
At each training iteration, the rollout engine generates responses for a batch of 32 prompts, sampling 5 responses per prompt with temperature 0.6 and top-$p$ 0.95. The maximum sequence length is set to 32,768 tokens. The teacher model then performs a forward pass over these rollout sequences, and the resulting logits are used to supervise the student in the subsequent training phase.
The student is trained with a reverse-KL distillation objective, with a global batch size of 16 and a per-device micro-batch size of 1. We use the AdamW optimizer with a peak learning rate of $2\times10^{-6}$ and a linear warmup over the first 5\% of training steps. Training runs for 10 epochs with gradient checkpointing enabled throughout.
A key memory challenge in on-policy distillation arises from materialising student and teacher logit tensors simultaneously: for a 32k-token response and a vocabulary of approximately 152k tokens, storing both logit matrices together with their gradients requires a number of GPU memory. To address this, top-$k$ OPD \citep{fu2026revisiting} calculates the KL divergence using only the top-32 tokens in the vocabulary. Experimental results demonstrate that this approach significantly reduces the computational burden without compromising the overall performance.

\section{More Experimental Results}

In this section, we provide more experimental results and analysis for \baby.

\begin{figure}[t]
  \centering
  \includegraphics[width=1.0\textwidth]{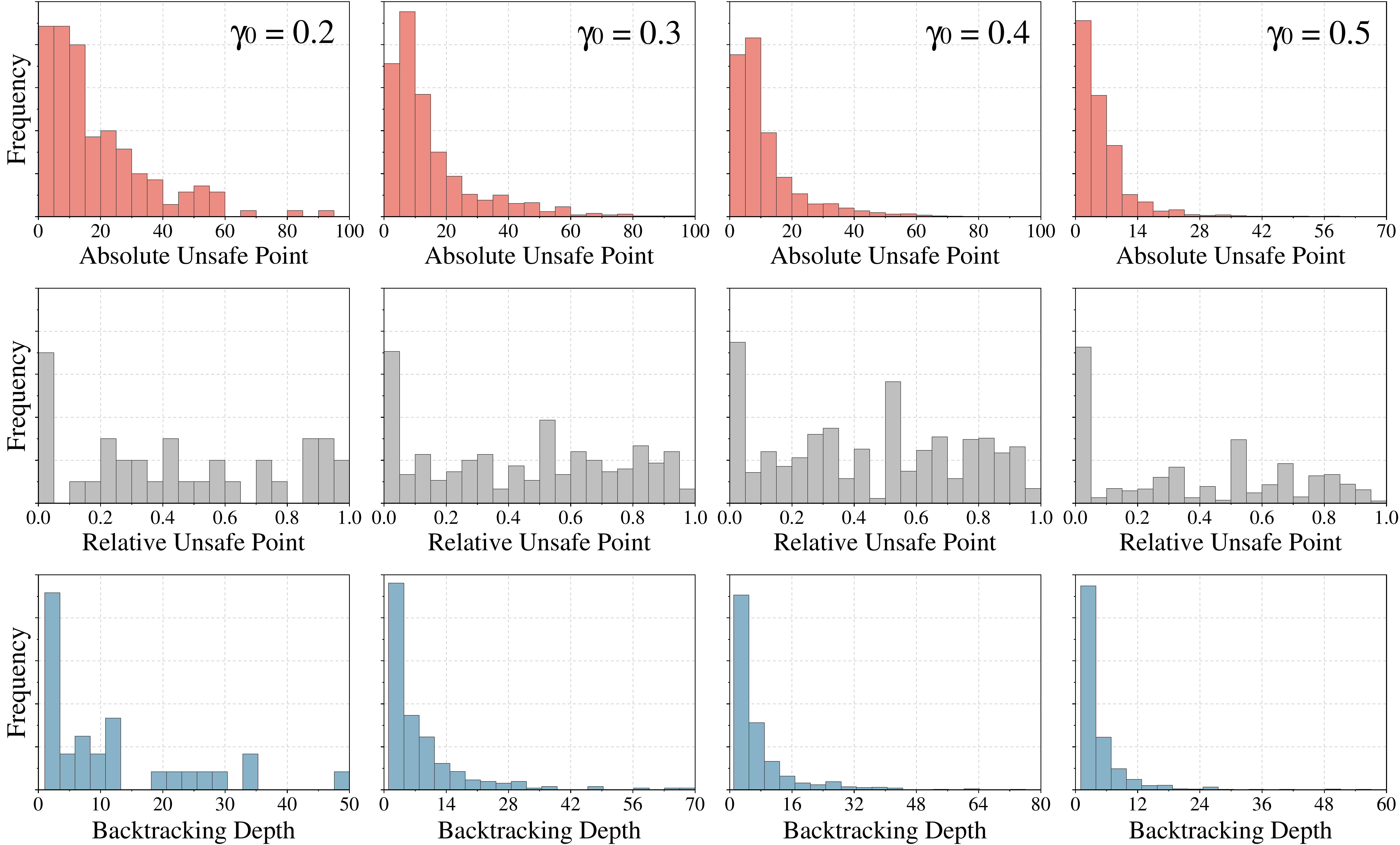}
  \vspace{-5pt}
    \caption{Distributions of absolute unsafe points (top row), relative unsafe points (middle row), and backtracking depths (bottom row) across varying base safety thresholds $\gamma_0 \in \{0.2, 0.3, 0.4, 0.5\}$.}
    \label{fig:backtracking_dynamics}
\end{figure}

\subsection{Experiments on More Student Models}

To further substantiate the robustness and generalizability of our proposed framework, Table~\ref{table:app.more_student} provides comprehensive experimental results across a broader spectrum of student models, utilizing Qwen3-32B as the teacher. Specifically, we evaluate \baby on diverse model scales and architectures, including Qwen3-1.7B, Llama3.2-3B-Instruct, and Qwen2.5-7B-Instruct. Across all three student models, \baby demonstrates consistent and remarkable superiority over both standard off-policy (\eg SFT, RFT) and on-policy (\eg ImitKD, SKD) baselines. For instance, on the extremely lightweight Qwen3-1.7B model, \baby dramatically improves the AIME24 Pass@1 from 9.58\% (SFT) to 22.08\%, and elevates the most challenging MATH500 Level 5 accuracy to 58.21\%, significantly eclipsing all baseline methods. Furthermore, the results on Llama3.2-3B highlight the efficacy of \baby in cross-architecture distillation scenarios. Despite the inherent stylistic and distribution discrepancies between the Qwen teacher and the Llama student, our method successfully bridges this gap, consistently pushing the performance upper bound. Notably, we observe that heuristic-based on-policy methods like SKD exhibit severe sensitivity to their manual thresholds $\beta$, with their optimal configurations fluctuating heavily depending on the specific student architecture and evaluation metric. In contrast, the entropy-driven adaptive boundary in \baby seamlessly adjusts to the student's intrinsic distribution, systematically yielding optimal performance without requiring exhaustive hyperparameter tuning. Beyond mathematical reasoning, \baby also delivers robust enhancements on general instruction following (IFEval) and advanced scientific knowledge (GPQA) tasks. This comprehensively verifies that the self-correction capabilities and error-recovery mechanisms instilled by our trajectory stitching and backtracking are domain-agnostic, ultimately fostering a more versatile and reliable reasoning policy across varying base models.

\subsection{Analysis of Backtracking Dynamics}
\label{subsec:backtracking_dynamics}

To better understand the intervention behavior of \baby, we visualize the distributions of detected unsafe points and the corresponding backtracking depths across varying base thresholds ($\gamma_0 \in \{0.2, 0.3, 0.4, 0.5\}$). As illustrated in Fig.~\ref{fig:backtracking_dynamics}, while the absolute unsafe points (top row) tend to cluster in the earlier steps of generation, the relative unsafe points (middle row) exhibit a remarkably uniform distribution across the entire lifespan of the trajectories. This uniformity underscores that logical deviations are not strictly confined to the initial planning phases but can emerge dynamically at any stage of complex reasoning, thereby necessitating the continuous, step-by-step monitoring implemented in our framework. More importantly, the distribution of backtracking depths (bottom row) provides compelling empirical validation for our credit assignment mechanism. Across all threshold settings, a substantial proportion of interventions require rewinding by multiple steps (depth $> 1$) rather than merely correcting the immediately preceding step. This phenomenon corroborates our core hypothesis regarding the delayed manifestation of reasoning errors: a critical boundary breach detected at step $t$ is frequently rooted in a flawed premise established several steps prior. By successfully tracing the temporal difference error back to the genuine safe bifurcation point, the credit assignment strategy ensures that the teacher model intervenes exclusively within a pristine, uncorrupted context, thereby fundamentally circumventing the reversed exposure bias.

\begin{figure}[t]
  \centering
  \includegraphics[width=1.0\textwidth]{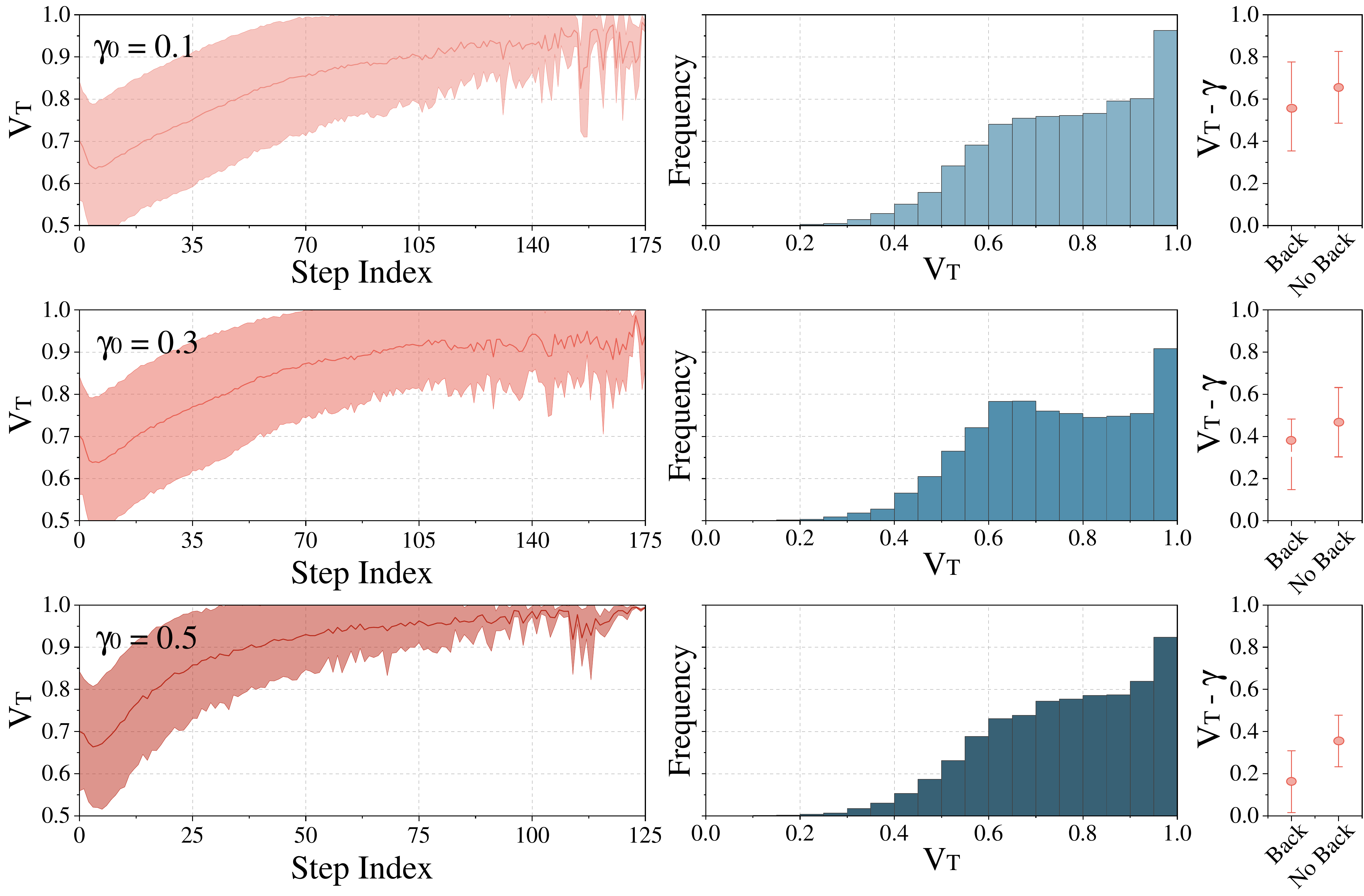}
  \vspace{-5pt}
    \caption{Evolution of the step-level value $V_T$ across reasoning steps (left), the overall distribution of $V_T$ (center), and the safety margins ($V_T - \gamma$) for backtracked versus non-backtracked trajectories (right), evaluated under different base thresholds $\gamma_0 \in \{0.1, 0.3, 0.5\}$.}
    \label{fig:value_dynamics}
\vspace{-5pt}
\end{figure}

\subsection{Dynamics of Teacher Confidence and Safety Margins}
\label{subsec:value_dynamics}

To further investigate how the safety boundary influences the reasoning process, we analyze the dynamics of the teacher's step-level value $V_T$ across different base thresholds ($\gamma_0 \in \{0.1, 0.3, 0.5\}$). As depicted in the left column of Fig.~\ref{fig:value_dynamics}, $V_T$ generally exhibits an upward trend as the step index increases, indicating that logical certainty solidifies as the trajectory approaches the final answer. Notably, as $\gamma_0$ increases, $V_T$ converges to 1.0 much more rapidly. This demonstrates that a stricter boundary compellingly restricts the student to the teacher's high-confidence manifold, accelerating logical convergence but concurrently compressing the exploratory space. This observation is corroborated by the $V_T$ distributions (center column), which become increasingly heavily skewed toward 1.0 at higher $\gamma_0$ values. Furthermore, we evaluate the safety margin, defined as $V_T - \gamma$, to compare trajectories that triggered backtracking against those that did not (right column). As $\gamma_0$ increases, the discriminative gap in the safety margin between the ``Back'' and ``No Back'' data progressively widens. This indicates that a tighter boundary enhances the framework's sensitivity, more decisively distinguishing fundamentally flawed deviations from valid logical progressions. However, synthesizing this with our main results reveals a crucial trade-off: while a larger $\gamma_0$ provides sharper error detection and faster confidence convergence, it prematurely truncates acceptable sub-optimal explorations, thereby hindering the mitigation of exposure bias. Consequently, a moderate boundary (\eg $\gamma_0=0.3$) is vital to balance discriminative error detection with adequate distributional coverage.

\begin{figure}[t]
  \centering
  \includegraphics[width=1.0\textwidth]{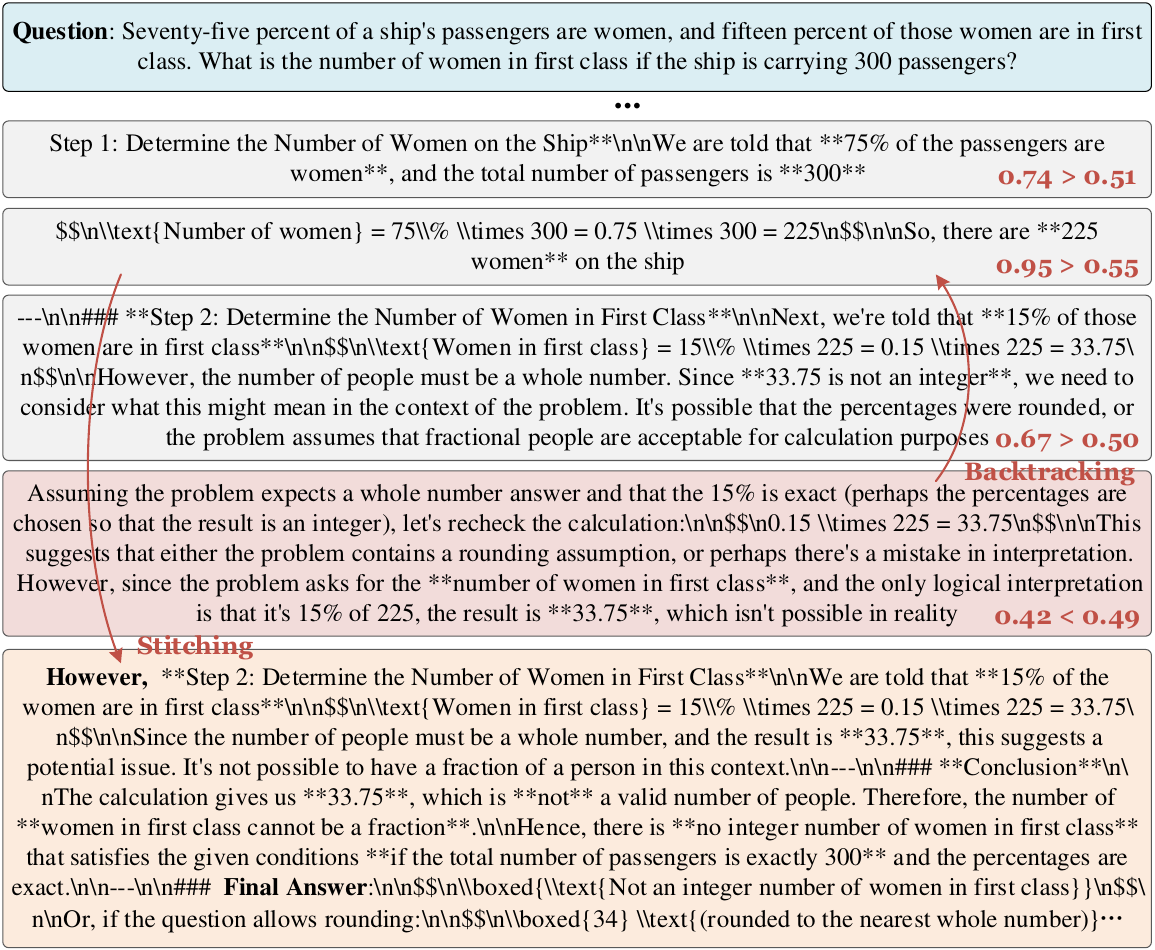}
  \vspace{-5pt}
    \caption{Qualitative case study of the \baby framework on an arithmetic word problem. The student's illogical ramble upon encountering a non-integer human count drops below the safety boundary ($0.42 < 0.49$), prompting a backtrack and a teacher-guided correction stitching.}
\label{fig:case_1}
\end{figure}

\begin{figure}[t]
  \centering
  \includegraphics[width=1.0\textwidth]{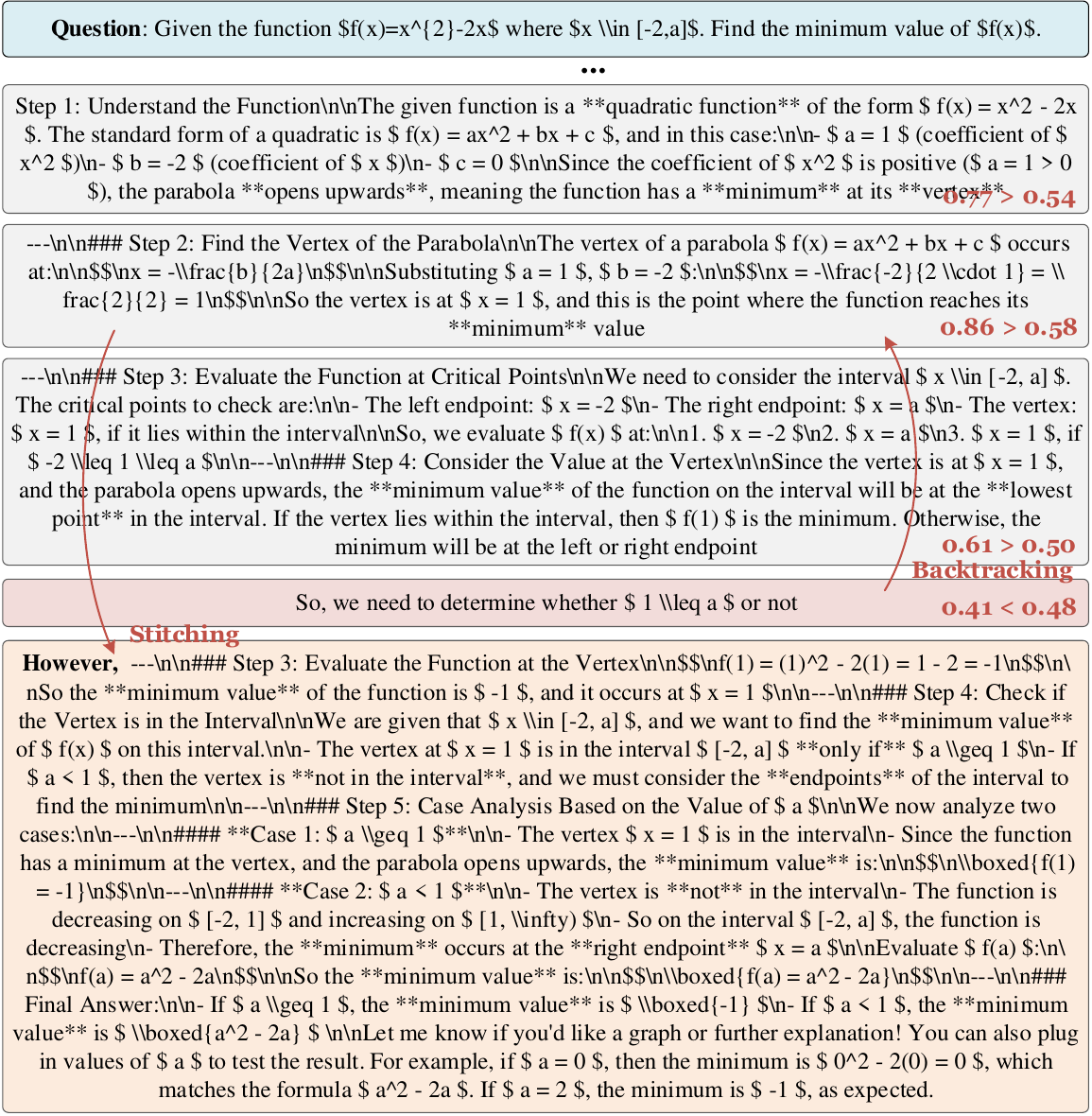}
  \vspace{-5pt}
    \caption{Qualitative case study of the \baby framework on a parameterized quadratic optimization problem. A premature logical leap regarding boundary conditions triggers intervention ($0.41 < 0.48$), seamlessly pivoting the trajectory into a rigorous case-by-case analysis.}
\label{fig:case_2}
\end{figure}

\subsection{Qualitative Case Studies}
\label{subsec:case_study}

To intuitively demonstrate how the \baby framework operates in practice, we present two qualitative case studies. In the first case, illustrated in Fig.~\ref{fig:case_1}, the student is tasked with an arithmetic word problem. The student correctly calculates an intermediate result of 33.75 women; however, because a human count must logically be an integer, the student subsequently spirals into confusion, generating an uncertain and rambling interpretation that violates the safety boundary ($0.42 < 0.49$). Instead of discarding the entire attempt or allowing the error to cascade, our framework backtracks to the precise point before the logical collapse. It then stitches the flawed exploration with a teacher-generated correction that rigorously formalizes the impossibility of the scenario. In the second case, presented in Fig.~\ref{fig:case_2}, the student attempts to find the minimum of a quadratic function within a parameterized interval $[-2, a]$. After successfully calculating the vertex, the student makes a premature and incomplete logical leap regarding the boundary condition, causing the value score to drop below the threshold ($0.41 < 0.48$). \baby traces the temporal difference error back to the vertex calculation and injects a comprehensive case-by-case analysis. By seamlessly stitching this pristine, teacher-guided correction to the student's flawed attempt using the semantic token ``\textbf{However}'', the resulting trajectory explicitly teaches the student how to recognize its own premature conclusions and pivot toward a more systematic analytical approach. Together, these cases illustrate how \baby effectively preserves the student's natural reasoning style while providing timely, high-quality interventions precisely when the intrinsic logic breaks down.

\section{Limitations}
\label{sec:limitations}

Although \baby fundamentally mitigates the dual exposure biases and significantly enhances the student model's reasoning capabilities, it inevitably introduces additional computational overhead during the data synthesis phase. Specifically, continuously monitoring the student's step-wise generation, calculating the teacher's conditional likelihoods, and estimating predictive entropy for the adaptive boundary necessitate frequent, synchronous interactions between the student and teacher models. This renders the offline trajectory generation process more computationally intensive compared to traditional off-policy methods. However, it is crucial to emphasize that this computational cost is strictly confined to the offline training preparation stage; once fine-tuned, the student model's inference efficiency and deployment costs remain completely unaffected. Given the substantial performance gains on complex reasoning tasks and the acquisition of intrinsic self-correction abilities, trading offline computational resources for high-quality, bias-mitigated distillation data represents a highly justifiable and practical compromise. Meanwhile, we provide a detailed formal and experimental complexity analysis in Appendix~\ref{subsec:complexity}.

\section{LLM Usage} \label{sec:llmusage}

LLMs were utilized solely for grammatical refinement and linguistic polishing of the manuscript; the core ideas, methodology, and results were developed independently by the authors.



\end{document}